\documentclass[11pt,a4paper]{article}
\usepackage{times,latexsym}
\usepackage{url}
\usepackage[T1]{fontenc}

\usepackage{amsmath}
\usepackage{graphicx}
\usepackage{algorithm}
\usepackage{algpseudocode}
\usepackage{xcolor}
\usepackage{pgfplots}
\usepackage{tikz}
\usepackage{hyperref}
\usepackage{amssymb}
\usepackage{booktabs, multirow}
\usepackage{enumitem}
\usepackage{tcolorbox}
\tcbuselibrary{breakable}
\usepackage{colortbl}
\usepackage{tcolorbox}

\usetikzlibrary{positioning, arrows.meta, calc, backgrounds}

\usepackage{amsfonts}
\usepackage[capitalise]{cleveref}

\definecolor{bblue}{HTML}{4F81BD}
\definecolor{rred}{HTML}{C0504D}
\definecolor{ggreen}{HTML}{9BBB59}
\definecolor{ppurple}{HTML}{9F4C7C}
\definecolor{oorange}{HTML}{F79646}
\definecolor{dgreen}{HTML}{003A00}
\definecolor{deeppurple}{RGB}{76,87,133}
\definecolor{dutch_white}{RGB}{234, 222, 189}
\definecolor{cool_gray}{RGB}{159, 167, 200}
\definecolor{cornell_red}{RGB}{173, 5, 13}

\definecolor{darkblue}{rgb}{0, 0, 0.5}
\hypersetup{colorlinks=true, citecolor=darkblue, linkcolor=darkblue, urlcolor=darkblue}

\usepackage[acceptedWithA]{tacl2021v1}

\usepackage{xspace,mfirstuc,tabulary}

\newif\iftaclinstructions
\taclinstructionsfalse 
\iftaclinstructions
\renewcommand{\confidential}{}
\renewcommand{\anonsubtext}{(No author info supplied here, for consistency with
TACL-submission anonymization requirements)}
\newcommand{\instr}
\fi

\iftaclpubformat 

\else

\fi

\makeatletter
\renewcommand{\sectionautorefname}{\S\@gobble}
\renewcommand{\subsectionautorefname}{\S\@gobble}
\makeatother

\title{Lightweight Latent Reasoning for Narrative Tasks}

\author{
  Alexander Gurung$^\diamond$ 
  \and
  Esmeralda S. Whitammer$^{\diamond\dagger}$
  \and
  Mirella Lapata$^\diamond$ 
  \\
  \ \\
  $^\diamond$School of Informatics, University of Edinburgh\\$^\dagger$CIFAR Fellow\\
  \texttt{\{alex.gurung,esmeralda.whitammer\}@ed.ac.uk}, \ \texttt{mlap@inf.ed.ac.uk}
}

\date{}

\begin{document}
\maketitle
\begin{abstract}
    Large language models (LLMs) tackle complex tasks by generating long chains of thought or
    ``reasoning traces'' that act as latent variables in the generation of an output given a query. A model's ability to generate such traces can be optimized with reinforcement learning (RL) to improve their utility in predicting an answer. This optimization comes at a high computational cost, especially for narrative-related tasks that involve retrieving and processing many tokens.
    To this end, we propose LiteReason, a latent reasoning method that can be interleaved with standard token sampling and easily combined with RL techniques. 
    LiteReason employs a lightweight Reasoning Projector module, trained to produce continuous latent tokens that help the model `skip' reasoning steps. During RL, the policy model decides when to activate the projector, switching between latent and discrete reasoning as needed. Experimental results on plot hole detection and book chapter generation show that our method 
    outperforms  latent reasoning baselines and comes close to matching non-latent RL training, while reducing final reasoning length by 77--92\%. Overall, LiteReason guides RL training to a more efficient part of the performance-computation tradeoff curve.\footnote{Code is available \href{https://github.com/Alex-Gurung/LiteReason}{here}.}
\end{abstract}

\section{Introduction}
\begin{figure*}[h]
\begin{center}

\input{inference-figure}
\end{center}
\vspace{-.5cm}
\caption{High-level diagram of LiteReason. Discrete sampling (via the
  LM head) is performed as normal, selecting a token (e.g.,~$x_1$) and passing its
  corresponding discrete token embedding, until we encounter special
  `implicit-thought' tags represented here by {``$\langle$bot$\rangle$''} (Beginning Of Thought). This LLM-generated tag also contains a number representing `thought complexity'. We then switch to \textbf{latent reasoning mode}
  and use the Reasoning Projector to directly predict continuous token
  embeddings (e.g., $e_0$), which are fed back through the model. This is repeated the `thought-complexity' number of times, before switching
  back to discrete sampling. 
  We can switch between discrete and reasoning mode multiple times before producing the final answer with discrete sampling. When training the Reasoning Projector we
  1) randomly replace reasoning steps with the  implicit thought tags
  and 2) freeze the rest of the LLM and apply a cross-entropy loss on
  the remaining reasoning steps (discrete
  tokens).
  }
\label{fig:diagram}
\end{figure*}
 
Reasoning has become a popular paradigm for improving LLM performance across tasks. Originally introduced through prompting methods such as Chain-of-Thought \citep{wei_chain--thought_2022}, which has models generate intermediate tokens (traces) before the final answer, reasoning has since evolved toward \mbox{RL-based} approaches that refine the ability of models to generate high-quality reasoning traces. Tasks in domains where answers can be verified, like math and coding, have seen significant improvement from RL training using the verifier as a reward signal \citep{shao_deepseekmath_2024, lambert2025tulu}. Recent work has begun extending RL methods to non-verifiable domains like story generation and understanding \citep{ahuja_finding_2025, gurung_learning_2025}.

However, performance gains from RL often come at increased cost during inference and training as reasoning traces become longer
\citep{shen_long_2025}.  This has motivated efforts to improve the \emph{efficiency} of reasoning, aiming to speed up both RL and inference-time generation \citep{sui_stop_2025}. 
One such effort is \textbf{latent reasoning}, which produces traces formed of \textit{continuous} token embeddings instead of or in addition to discrete tokens.
The justification for these latent reasoning methods is twofold: firstly, many tokens are perceived not to contain information useful for predicting the answer; 
secondly, human reasoning is often done implicitly without verbalization.  Keeping thoughts in a non-token space affords   flexibility, allowing us to retain more information than lossily decoding to tokens.
For example, \citet{hao_training_2025} argue that continuous embeddings can represent a mixture of reasoning steps, which effectively approximates performing several reasoning rollouts in parallel.

In this work, we introduce LiteReason, a latent reasoning algorithm designed for narrative understanding and generation tasks, where both inputs and outputs often span thousands of tokens. 
LiteReason adds a Reasoning Projector head to a base LLM, allowing the generation both of text tokens via discrete sampling and of continuous embeddings via \textbf{latent reasoning mode}.
This mode is activated by the production of a special start-of-thought symbol. 
(See \autoref{fig:diagram} for a schematic illustration.) 
After pretraining the Reasoning Projector, we perform RL fine-tuning of the base model to encourage efficient reasoning with  latent tokens and  improve performance at a given task.

LiteReason is especially well-suited to narrative tasks.
    In particular, we focus on detecting plot holes in stories (FlawedFictions; \citealt{ahuja_finding_2025}) and generating book chapters, referred to as Next Chapter Prediction (NCP; \citealt{gurung_learning_2025}). Both tasks require a variety of story-understanding capabilities, asking models to reason over \mbox{$>1$k} tokens of story context to respectively detect plot inconsistencies and plan the next chapter.
They also introduce some key difficulties when adapting existing latent reasoning approaches. Firstly, they lack large datasets of high-quality reasoning traces, and because default model performance is low, collecting such  datasets is difficult. Secondly, they require high levels of adaptability, as the input space of stories is very diverse. These challenges guide the design of LiteReason, which balances language model exploration with lower inference costs.

We evaluate LiteReason on both verifiable (Flawed Fictions) and non-verifiable  (NCP) tasks and compare against trained \citep{hao_training_2025, tan_think_2025} and training-free latent reasoning methods \citep{zhuang_text_2025, zhang_soft_2025}. We also include RL-trained models without latent reasoning as an upper bound for performance and a benchmark for computational costs. We find that LiteReason outperforms all latent-reasoning baselines and reaches 69--96\% of the performance gain of non-latent RL training while generating~\mbox{50--53}\% fewer tokens during training and generating~70\% fewer tokens during inference. Our core contributions and findings are:

\begin{itemize}[itemsep=0pt,  left=0pt]
    \item We introduce LiteReason, a new method that integrates latent reasoning with RL, relying only on the pretrained LLM (and no preexisting dataset of reasoning traces) and designed for low-resource narrative tasks. 
    \item Empirical results on Flawed Fictions (verifiable) and Next Chapter Prediction (non-verifiable) tasks show high performance relative to latent reasoning and RL baselines.   
    \item Further analysis shows our method results in highly efficient solutions and requires significantly fewer tokens during training and inference than traditional RL.
\end{itemize}

\section{Related Work}

Initial work demonstrated that Chain-of-Thought prompting can enhance LLM capabilities by generating intermediate tokens before predicting the final answer \citep{wei_chain--thought_2022, kojima2022large}. Recent research has successfully applied RL to optimize these reasoning traces, particularly in domains like math and code \cite{shao_deepseekmath_2024,kimi_team_kimi_2025,deepseek-ai_deepseek-r1_2025,lambert2025tulu}.
However, these performance gains often come with increased inference costs as reasoning chains become longer \citep{shen_long_2025}. Furthermore, as RL training often requires generating many reasoning traces, the cost of training can also be significant. Consequently there has been interest in making reasoning more \textit{efficient}, exploring the tradeoff between reasoning length and performance \citep{sui_stop_2025}.

\paragraph{Reasoning in the Latent Space} Latent reasoning attempts to improve efficiency by performing reasoning in a continuous space instead of sampling discrete tokens. 
 Two primary strategies have been explored to achieve this: a \emph{training-free} approach, which modifies the sampling procedure without altering the model itself, and a \emph{training-based} approach, which teaches the model to construct latent representations of reasoning.

 Soft Thinking \citep{zhang_soft_2025} and Mixture of Inputs \citep{zhuang_text_2025} are both examples of training-free strategies. The former  uses a probability weighted mixture of the top-k token embeddings as the next input and adds an entropy-based stopping mechanism to switch back to discrete sampling. The latter similarly uses a mixture of embeddings, but the weights are determined via a Bayesian estimation method where the sampled token is treated as an observation. 
Although these methods are lightweight to implement, they typically only improve performance when the model's initial capabilities on the task are high. Both methods apply 27B+ models on datasets like AIME, GSM8K \citep{cobbe_training_2021}, GPQA Diamond \citep{rein_gpqa_2024} and LiveCodeBench \citep{jain_livecodebench_2024}, where pre-existing Chain-of-Thought performance is over 80\%.

One influential training-based method is COCONUT \citep{hao_training_2025}, which uses an LLM's final hidden state as the next input embedding and proposes a training curriculum that iteratively replaces more reasoning steps with these hidden states. Training the entire model to predict the remaining reasoning steps and final answer encourages it to store latent reasoning representations in this final hidden state. COCONUT is built on GPT-2 and trained to perform math and synthetic logical reasoning tasks on large datasets like GSM8K-Aug \citep{deng_explicit_2024}.
CoLaR \citep{tan_think_2025} proposes adding a latent head to the LLM that predicts token embeddings to replace reasoning steps, supervised by the token embeddings of the replaced steps. After fine-tuning on a large reasoning database, CoLaR performs 
RL directly over its latent tokens. Similar to COCONUT, at test-time CoLaR produces a series of latent tokens followed by standard discrete sampling and an answer. CoLaR is scaled up to larger models (Llama 3.2 1-8B Instruct) and is also trained with GSM8K-Aug for math benchmarks.

Other approaches include CoT2~\citep{gozeten_continuous_2025}, which
trains \mbox{GPT-2} to produce token mixtures for logical reasoning,
similar to training-free methods; CODI~\citep{shen_codi_2025}, which
applies teacher-student distillation to improve performance on GSM8k
with GPT-2 and Llama~3.1~1B; and Token Assorted~\citep{su_token_2025},
which learns new token embeddings via a VQ-VAE for math and logical
reasoning with Llama~3.1~8B. During inference, these new, fixed token embeddings can be used as standard entries in the model's vocabulary. 

Our LiteReason approach takes inspiration from  these previous trained latent-reasoning methods; we predict continuous latent tokens with a latent-head similar to CoLaR, and support interleaved latent reasoning like Token Assorted. However, LiteReason differs in a few ways:

Firstly, we do not rely on a large dataset of reasoning traces like GSM8k-Aug (as such datasets do not exist for our tasks), instead exclusively using traces generated from the model itself.

Secondly, unlike CoLaR and Token Assorted where the placement of latent reasoning is fixed during training, we allow the model to learn when and how to best interleave latent and discrete reasoning during RL training. This form of RL-training still operates in the discrete token space, allowing it to be easily combined with other reward-shaping methods.

Finally, although many existing approaches compare against Chain-of-Thought prompting and a few  incorporate RL, to our knowledge none of them compare final performance against a non-latent RL-trained model. We believe this baseline to be a more realistic point of comparison for difficult tasks, so we include it in our experiments as an upper bound on performance and compute.

\paragraph{Narrative Tasks}

Although much reasoning work has focused on domains like math and coding, there is a large body of work applying LLMs to narrative understanding and generation tasks \citep{mostafazadeh_corpus_2016, fan_hierarchical_2018, huot_agents_2024,
yang_re3_2022, yang_doc_2023,
huot_agents_2024, xie_creating_2024, chhun_human_2022, karpinska_one_2024, sprague_musr_2023}. Numerous studies on story generation prompt models to perform specific sub-tasks such as  planning plot events or generating character details \citep{huot_agents_2024, yang_doc_2023, yang_re3_2022}, which can be viewed as a way to elicit specific reasoning styles.
  However, these tasks are often very difficult to adapt to the RL domain due to their lack of verifiable rewards. We focus in this work on two specific tasks that evaluate narrative-based  reasoning in LLMs, while still providing clear reward signals for policy-gradient methods.

\citet{ahuja_finding_2025} constructed the Flawed Fictions benchmark by automatically synthesizing plot holes in human written stories, and asking models to predict whether the given story contains a plot hole. This task requires a variety of cognitive abilities, including theory of mind, accurate state tracking, and commonsense reasoning. We focus on the verifiable binary prediction task (Yes/No answer to the question: does the story have a plot hole?), where initial performance is only slightly above 50\% (random).

\citet{gurung_learning_2025} introduced the difficult Next Chapter-Prediction (NCP) task, where LLMs are used to predict \textit{plans} for  the next chapter in a book based on information about its story, characters, and previous chapters. 
During RL, the quality of a plan or reasoning trace is evaluated by how much it improves the likelihood of reproducing the true next chapter. 
This non-verifiable task is especially challenging for existing latent reasoning methods, as the output (plans) are long and linguistically diverse.

\section{The LiteReason Framework}
\label{sec:methodology}

In this section, we introduce a method that integrates latent reasoning with reinforcement learning (RL) to improve performance while substantially reducing computational cost. 
This method is designed around key assumptions imposed by the nature of narrative analysis and generation tasks:

\begin{enumerate}
[wide, labelwidth=!, labelindent=0.25pt]
    \item We lack a large, high-quality dataset of reasoning traces. This limitation is especially common for long-form generation tasks or analysis tasks where the reasoning space is very large. This is also often true for tasks where RL is chosen \textit{because} exploration is needed to achieve good performance and distillation is impossible. We do, however, assume our tasks provide a reward function for evaluating a given response.
    \item We would like to retain standard language modeling capabilities, as our responses and reasoning should be highly diverse.
\item  We assume our model is capable of some reasoning, so we can initialize our method from reasoning traces taken from the model itself.
\end{enumerate}

Our only architectural addition to the base LLM is a small Reasoning Projector, composed of a small stack of MLP layers. Similar to previous latent-reasoning work (e.g., \citealt{hao_training_2025}), this projector takes in the last hidden state and predicts a continuous token-embedding vector. This module is only called during latent-reasoning; otherwise, the last hidden state is passed to the language model head and a discrete token is sampled like normal. We refer to these two pathways as \textbf{latent reasoning mode} and \textbf{discrete mode}, respectively. \autoref{fig:diagram} shows how these modes are interleaved throughout inference and training, and this process is described in more detail below.

Our goal is to train the  Reasoning Projector (via supervised fine-tuning; SFT) such that its `latent thoughts' help the LLM skip reasoning steps, and  the LLM (via RL) to both improve at the given task \emph{and} choose when to use the Reasoning Projector.


\subsection{Training LiteReason}

There are three stages to training LiteReason, described below: (1) Data Collection,  (2) SFT Initialization,  and (3) RL Training. Our inference procedure is shown in \autoref{fig:diagram}.

\paragraph{Data Collection}

We first wish to collect a small dataset of reasoning traces to initialize the Reasoning Projector to produce `latent thoughts' (token embeddings) useful for skipping reasoning steps. We construct the initial SFT dataset via rejection sampling, where we sample $n$~reasoning traces for every prompt in the training set and filter out the traces with non-positive reward (Accuracy for Flawed Fictions, Contrastive Improvement for NCP; see \autoref{sec:application} for details). We hypothesize that filtering out non-positive reward traces encourages our reasoning projector to learn more useful and correct reasoning patterns. 
We split these reasoning traces into steps by sentences and randomly replace a proportion~$s_r$ of them 
with \textbf{implicit thought tags}, which have the  structure:  

\begin{tcolorbox}
\centering
\!\!\!\!\footnotesize\texttt{<implicit\_thought>\#</implicit\_thought>}
\end{tcolorbox}
\noindent
where \# is an integer representing the number of steps to take in latent reasoning mode. At each step we pass the final hidden state to the Reasoning Projector and append the predicted token embedding to the sequence. These predicted `latent thoughts' are trained to maximize the likelihood of reasoning tokens. We set \# as a proportion $t_r$
of the number of tokens in the sentence being replaced. We expect these
hyper-parameters, $t_r$ (token-replacement ratio), $s_r$
(sentence-replacement ratio), and $n$ (sample number), to be optimized for new
model-task combinations.  
We tuned these values in preliminary
experiments to balance cost and the amount of information latent
tokens represent. The chosen values are reported in \autoref{sec:hyperparameterdetails}.

\paragraph{Supervised Fine-Tuning (SFT)}

Once we have a dataset of reasoning traces with steps randomly replaced
with implicit-thought tags, we aim to train our Reasoning Projector to
predict useful token embeddings for `skipping' reasoning steps.  We
freeze all model parameters except the projector and train using a
standard cross-entropy loss on the remaining reasoning sentences. This
encourages the projector to produce embeddings that `bridge the gap'
between explicit reasoning steps when skipping over the masked
reasoning step. As later latent reasoning tokens depend on earlier ones,
this requires backpropagation through a rollout of several steps of latent reasoning.

\paragraph{RL Training}
After initializing the Reasoning Projector, we improve performance on a given task via RL using the reward functions defined in \autoref{sec:application}. By incorporating latent reasoning into this process, we hope to reduce the computational cost required for improved performance. Although our RL training is agnostic to the choice of policy-gradient algorithm, we conducted our experiments using Group Relative Policy Optimization (GRPO; \citealt{shao_deepseekmath_2024}) to ensure consistency with previous work on latent reasoning and narrative generation \citep{gurung_learning_2025, tan_think_2025}. For each training instance, we sample a group of outputs $\{o_1, o_2, ... o_G\}$ from the old policy $\pi_{\text{old}}$, where $G$ is the group size (a hyper-parameter). The reward for each output is converted to an advantage via group normalization and used to update the policy.

During RL, we train the whole LLM, but do not compute policy gradients with respect to the prediction of latent thought tokens. In other words, we consider only the token sampling steps as `actions', some of which are conditioned on past latent thoughts. As a result, the Reasoning Projector is not updated during RL, although latent thoughts may change because the discrete token and latent thought predictors share the model body.
We hypothesize that during training our Reasoning Projector may fall out-of-date with the current reasoning traces produced by our model.

To alleviate this, we add an SFT step after every epoch of RL training that performs the same SFT procedure as described above, but on the reasoning traces produced during the epoch. In addition to the latent tokens produced by the model, we also randomly replace a proportion of sentences with latent tokens, just as in the SFT phase. We find that this further encourages the model to produce efficient reasoning. Future work could explore a joint-objective approach where the reasoning projector is updated continually throughout training instead of in stages. 

In experiments, we refer to the variant without these periodic SFT steps as `LiteReason w/o SFT'. 

\subsection{Inference Procedure}
During inference we default to \textbf{discrete mode}, but when
encountering implicit-thought tags we switch to \textbf{latent
  reasoning mode} (see \autoref{fig:diagram}). As mentioned
earlier, we delineate between discrete and latent reasoning modes via
\textbf{implicit thought tags}. Note that these tags are
not special tokens in the LLM's vocabulary, so instead of training the
model to encourage their use we can simply prompt the model. The model also predicts the number \#, deciding how many latent tokens are produced. Initially, 
usage of these tags is low, but we find that during early RL training
they become much more common. 

To detect these implicit thoughts we perform a small sweep across common string variants to give us a fixed list of accepted implicit thought token patterns, and look backward at each generated token to see if our most recent generation matches a pattern. If so, we attempt to parse the \# as an integer, and ignore invalid values.

During latent reasoning mode, for the given number (\#)~of steps we do not
sample a new token, but instead pass the last hidden state to the
Reasoning Projector and feed the output as the next token
embedding. After the last reasoning step we switch back to sampling
tokens as normal. Similar to other methods with latent reasoning heads
(e.g.,~CoLaR; \citealt{tan_think_2025}), this incurs a small
computational cost when producing latent tokens. In practice,
latent tokens are about 8\% more expensive than normal
tokens, but this cost is far outweighed by the savings in producing
fewer tokens.
\begin{table*}[t]
{\small
\resizebox{\textwidth}{!}{
\begin{tabular}{@{}ll@{}}
\toprule
\multicolumn{1}{@{}l}{\bf Dataset}  &\multicolumn{1}{l}{\bf Example Reasoning Step} \\
\hline
GSM8K-Aug \citep{cobbe_training_2021} & \textit{The helmet costs \$15x 2 = \$30.} \\
ProsQA \citep{hao_training_2025} & \textit{Every bompus is a wumpus.} \\
ProntoQA \citep{PrOntoQA}& \textit{Each vumpus is mean.} \\ \hline
Flawed Fictions &  \textit{The continuity
  error occurs because the story earlier establishes that the little girl} \\
  \citep{ahuja_finding_2025} &  \textit{was very poor and had no room to live in or bed to sleep in,
    but later it states that} \\
  & \textit{she returned to her small bed in the
  shelter with her newfound wealth.}\\ \hline
Next Chapter-Prediction  & \textit{<citation>Source A (Character
  Sheet: Rose) says X Rose is
  rebellious, disobedient,} \\
\citep{gurung_learning_2025} &  \textit{and has a sarcastic sense of
  humor.</citation>, therefore <reasoning>Rose will} \\
&  \textit{likely continue to
  challenge authority figures and express her opinions, possibly}\\
& \textit{provoking Miss Wellwood and leading to a confrontation.</reasoning>}
\\  \bottomrule
\end{tabular}
}
\caption{Example reasoning steps taken from reasoning traces in previously tested datasets
  (top) compared to our tasks (below). Our tasks involve significantly
  more varied and complex reasoning steps, which we hypothesize is a
  more difficult and realistic setting for latent reasoning. Steps for
  Flawed Fictions and NCP were taken from Qwen 2.5 7B-Instruct. For Flawed Fictions the model compares two story events implicitly, and for NCP the model explicitly cites and reasons over the given Story Information.} 
\label{tab:example_reasoning_steps}
}
\end{table*}

\section{Application to Narrative Tasks}
\label{sec:application}

\begin{table}[t]
{\small
\begin{center}
\begin{tabular}{lclr}
\toprule
\multicolumn{1}{@{}c}{\bf Dataset}  & \multicolumn{1}{c}{\bf \# Examples} & \multicolumn{1}{c}{\bf \# Tokens} & \multicolumn{1}{c}{\bf Source} \\
\midrule
Flawed & \ 414 & $\approx$ 900 & Project \\
Fictions & & & Gutenberg \\
NCP & \ \hspace{-1.5ex}1,347 & $\approx$ 6.5k & Recent \\
& & & Books \\
\bottomrule
\end{tabular}
\end{center}
\caption{Size and provenance for the Flawed Fictions \citep{ahuja_finding_2025} and Next Chapter Prediction \citep{gurung_learning_2025} datasets. \# Tokens describes the mean number of \textit{input} tokens for each example. \# Examples is the total number of datapoints in the dataset, we use the prescribed train/test/val split for NCP and a random 70/15/15 split for Flawed Fictions.}
\label{tab:dataset_statistics}
}
\end{table}

In this section we briefly present the Flawed Fictions and
Next Chapter Prediction tasks used to evaluate our method. We
describe how these tasks are originally formulated and how a vanilla
reasoning model (without latent thoughts) is trained via RL.  As
described below, both tasks require complex story-driven reasoning
that is semantically and stylistically distinct than traditional
tasks used by latent-reasoning
methods. \autoref{tab:example_reasoning_steps} displays example
reasoning steps from these tasks, and \autoref{tab:dataset_statistics} describes our datasets.

\subsection{Flawed Fictions}
\label{sec:ff}

Flawed Fictions is a benchmark proposed by \citet{ahuja_finding_2025}
that tests a variety of narrative-based reasoning abilities like
theory-of-mind and state-tracking. The dataset is constructed on short stories from Project Gutenberg by
FlawedFictionsMaker, an algorithm that controllably
\textit{induces} plot holes and continuity errors using LLMs, and filtered with human annotators (see \autoref{tab:dataset_statistics}  for dataset statistics).

Models are given a story and asked to
respond Yes or No if the story contains a plot hole. Open-source models like Qwen2.5-32B score around~53\%, only
slightly better than random~(50\%). Even large closed-source models
like o3-mini only achieve 63\%~accuracy. 
Although the benchmark can be extended to longer stories or specific-line plot hole detection, in this work  we focus
on the original binary prediction task, reporting Accuracy as the
performance metric.

Our reward for RL is a simple binary function in the style of Reinforcement Learning with Verifiable Rewards (RLVR; \citealt{lambert2025tulu}, and we use a slight prompt change that we found improved RL training performance. More details and prompts are presented in \autoref{sec:hyperparameterdetails}.

\noindent\[
R_{\text{flawed}} = \begin{cases} 
      0 & \text{pred } \neq \text{answer} \\
      1 & \text{pred } = \text{answer}
   \end{cases}
\]


\subsection{Next Chapter-Prediction}
\label{sec:ncp}

Next Chapter-Prediction, as proposed by \citet{gurung_learning_2025},
involves first constructing a detailed plan for the next chapter of a
book, and then generating the chapter itself based on that plan.  This
task is particularly challenging, requiring  reasoning over 10k+
tokens and producing rich long-form answers (the plans are around 100
tokens). To facilitate model training, \citet{gurung_learning_2025} collect a dataset 
of 30 books published in or after 2024 (ranging from 67k to 214k tokens, with a mean of 139k),  which we also use in experiments (see \autoref{tab:dataset_statistics} for dataset statistics). 

More formally, at a given chapter index~$i$, a \emph{reasoning model}
$\pi_{\theta}^{\mathcal{R}}$ reasons over Story Information~$SI_i$ and
predicts a plan $\hat{p}$ for the next chapter. $SI_i$~contains a
global story sketch, a plot summary of previous chapters, character
sheets from previous chapters, the immediately preceding chapter, and
a high-level synopsis of the next chapter. A \textit{story-generator
  model} $\pi^{\mathcal{G}}$ takes this plan and the
$SI_i$ and predicts the next chapter~$\hat{c}_{i+1}$:

\hspace{-1cm}\begin{minipage}{.5\linewidth}
\vspace{-.1in}
\begin{equation}
  \hat{p} \leftarrow \pi^{\mathcal{R}}_\theta (SI_i)
\end{equation}
\end{minipage}\hfill
\begin{minipage}{.5\linewidth}
\vspace{-.1in}
\begin{equation}
  \hat{c}_{i+1} \leftarrow
      \pi^{\mathcal{G}} (SI_i,\hat{p})
\end{equation}
\end{minipage}
\vspace{0.1in}

\noindent
Both reasoning and generator models are initialized with the same base
model (Qwen 2.5 7B-Instruct) but only the reasoning
model is trained. 

To learn useful reasoning steps and optimize the plan, which requires
extensive reasoning over the Story Information and a long-form answer,
\citet{gurung_learning_2025} propose \mbox{VR-CLI} (Verifiable Rewards
via Completion Likelihood Improvement), a proxy reward objective that improves the likelihood of generating the true
next chapter, conditioned on both the Story Information and the plan. $I$ refers to the \textit{Improvement} in perplexity ($\mathcal{P}$) of the chapter given the plan.

\vspace{0.1in}
\noindent
\begin{equation*}
\begin{split}
I_{\pi^{\mathcal{G}}}(x, y, a) &= [\frac{\mathcal{P}_{\pi^{\mathcal{G}}}(y|x) - \mathcal{P}_{\pi^{\mathcal{G}}}(y|x,a)} {\mathcal{P}_{\pi^{\mathcal{G}}}(y|x)}] \times 100
\end{split}
\end{equation*}


We found significant repetitions in reasoning traces using this objective, signaling a form of mode-collapse where the policy model learns to generate generic writing advice (e.g. `the chapter is interesting') instead of specific claims/plans for the next chapter.

We extend VR-CLI to include a \textbf{contrastive term}
that negatively weights the likelihood of a randomly chosen
chapter from another book:
\begin{align*}
    R_\text{contr} &= I(SI_i, c_{i+1}, \hat{p}) - \gamma I(SI_i, c^{r}, \hat{p}),
\end{align*}
which is a stochastic reward depending on the uniform random choice of the other chapter $c^r$. The contrastive term improves diversity in the final plans and
reasoning traces (see \autoref{tab:diversity_contrastive}), which we believe is a
better test case for LiteReason. We use this reward during training
and as our primary automated metric for evaluating NCP results. To
avoid confusion with the original improvement term, we refer to our reward $R_\text{contr}$ as \textbf{Contrastive Improvement}. We present details in \autoref{sec:extending_vrcli}
and compare example completions in \autoref{tab:compare_plans_contrastive}.

\begin{table}[t]
{\small
\begin{center}
\begin{tabular}{@{}ll@{~~}ll@{~~}l@{}}
\toprule
\multicolumn{1}{@{}l}{\bf Reward}  &\multicolumn{2}{l}{\bf Reasoning TTR} & \multicolumn{2}{c}{\bf Final Plan TTR} \\
\midrule
Default & 0.072 & & 0.046 &  \\
Contrastive & 0.087 &  (\textcolor{dgreen}{+20.8\%}) & 0.136 & (\textcolor{dgreen}{+200.2\%}) \\
\bottomrule
\end{tabular}
\end{center}
\caption{Comparing the Type-Token Ratio (TTR) of the reasoning and
  final plans on the NCP task, with models trained using the default
  objective (VR-CLI) and our Contrastive objective (Contrastive
  VR-CLI). Although we observe only a slight increase in reasoning diversity,
  we see a significant increase in final plan diversity. As shown in
  the examples in \autoref{tab:compare_plans_contrastive}, 
  our reasoning and final plans are more specific to the current
  chapter.}
\label{tab:diversity_contrastive}
}
\end{table}

\section{Experimental Setting}
\label{sec:experimental_setting}

We perform the main experiments with Qwen 2.5-7B Instruct \citep{qwen_qwen25_2024}, as it has shown good performance on the
NCP task \citep{gurung_learning_2025} and is amenable to RL training on the Flawed Fictions dataset (GRPO training leads to 
a large increase in performance, from $57\%$ to $89\%$). We call this
GRPO baseline RL-Trained, as described in
\autoref{ssec:baselines}. 
We study other models in \autoref{sec:other_models}.

\subsection{Testing LiteReason Variants}

\begin{table*}[t]

\resizebox{\textwidth}{!}{
{\small
\begin{tabular}{@{}ll@{~~}l@{~~}r@{~~}lr@{~~}l@{~~}l@{~~}l@{}}
\toprule
 & \multicolumn{4}{c}{\bf Flawed Fictions} & \multicolumn{4}{c}{\bf NCP} \\
 \multicolumn{1}{@{}l}{\textbf{Method}}  & \multicolumn{2}{c}{\textbf{Accuracy (\%)}} &  \multicolumn{2}{c}{\textbf{Avg \# Tokens}} & \multicolumn{2}{c}{\textbf{Contra Improve (\%)}} & \multicolumn{2}{c}{\textbf{Avg \# Tokens}} \\
\midrule
Qwen2.5-7B & 57.26  {\tiny $\pm$ 1.58} &  & 400.04  {\tiny $\pm$ 6.50} & & 0.067  {\tiny $\pm$ 0.001} &  & 831.85  {\tiny $\pm$ 8.26}  & \\
RL-Trained & 88.71 {\tiny $\pm$ 0.00} & (\textcolor{dgreen}{+31.45\%}) & 114.53 {\tiny $\pm$ 2.01} &(\textcolor{dgreen}{-71.37\%}) & 0.666 {\tiny $\pm 0.01$} &(\textcolor{dgreen}{+894.03\%}) & 721.33 {\tiny $\pm 6.01$} &(\textcolor{dgreen}{-15.97\%}) \\
\arrayrulecolor{gray} \hline
LiteReason &  87.42 {\tiny $\pm  0.47$} & (\textcolor{dgreen}{+30.16\%}) & 34.01 {\tiny $\pm 0.38$} & (\textcolor{dgreen}{-91.50\%}) & 0.478 {\tiny $\pm 0.01$} &(\textcolor{dgreen}{+613.43\%}) & 193.11 {\tiny $\pm 2.02$} & (\textcolor{dgreen}{-77.50\%}) \\
~~~w/o SFT & 85.48 {\tiny $\pm 0.00$} & (\textcolor{dgreen}{+28.22\%}) & 8.26 {\tiny $\pm 0.17$} & (\textcolor{dgreen}{-97.94\%}) & 0.560 {\tiny $\pm 0.01$} &(\textcolor{dgreen}{+735.82\%}) & 622.23 {\tiny $\pm 1.82$} & (\textcolor{dgreen}{-27.51\%}) \\
~~~w/o RP & 87.58 {\tiny $\pm 0.76$} &  (\textcolor{dgreen}{+30.32\%}) & 260.81 {\tiny $\pm 3.68$} & (\textcolor{dgreen}{-34.80\%}) & 0.449 {\tiny $\pm 0.01$} & (\textcolor{dgreen}{+570.15\%}) & 983.49 {\tiny $\pm 5.30$} & (\textcolor{purple}{+14.57\%}) \\
\bottomrule
\end{tabular}
    }}
\caption{We validate LiteReason design variants on the Flawed Fictions (left) and Next Chapter Prediction (right). All variants are trained using GRPO with the same hyper-parameters described in \autoref{sec:methodology}. All variants drastically improve over the untrained baseline, but the full LiteReason variant best balances high performance with significant token savings. Scores are averages across the test set, with $\pm \text{SEM}$ denoting the standard error of the mean. Percent differences are relative to the default Qwen 2.5-7B model (first row). Accuracy percent differences are $\text{method} - \text{default}$. All other columns are unbounded values, so we compute percent difference as $\frac{\text{method} - \text{default}}{\text{default}}$. Contra Improve refers to our Contrastive Improvement reward metric. RP abbreviates Reasoning Projector. 
}
\label{tab:name_variant_both_tasks}
\end{table*}

We test each component of LiteReason through experiments on Flawed Fictions and NCP. We use the following LiteReason instantiations which are all RL-trained, but vary in their use of SFT or Reasoning Projector. We use the following hyper-parameters (\autoref{sec:methodology}) when applicable: $t_r=0.2, s_r=(10\%,25\%)$, and $n=5$, where $s_r=(10\%,25\%)$ means a naive equal mixture of datapoints with $s_r=10\%$ and $s_r= 25\%$, which we found worked well even without a curriculum. Higher $t_r$ values increase expressivity per latent-thought, but increase training time and cost due to dependence of each latent embedding on the previous. 

\paragraph{LiteReason} Our standard method, where we periodically (once an epoch) run a small SFT training step using the trajectories generated during that epoch. This adds some computational cost, but updates the reasoning-projector with the current model's reasoning. We also use an implicit-thought prompt (\cref{tab:ncp_prompts,tab:flawed_fiction_prompts}) that encourages model usage of the implicit thought tags.

\paragraph{LiteReason w/o SFT} This version of LiteReason keeps the implicit prompt and initialized Reasoning Projector that uses implicit thought tags 
to know when to pass new reasoning embeddings to the model. However, this variant does not perform SFT during RL to update the Reasoning Projector. 


\paragraph{LiteReason w/o Reasoning Projector} Finally, we evaluate performance post RL without the Reasoning Projector (RP), but with the implicit thought prompt. This method does not use the latent reasoning, and tests if the model performs differently if we prompt for skipping reasoning steps. 

\subsection{Comparison Methods}
\label{ssec:baselines}

Drawing from the recent  latent reasoning literature, we select a representative set of previous methods to test the design assumptions outlined in~\autoref{sec:methodology}. We evaluate these methods on both the Flawed Fictions and NCP tasks.

\paragraph{Training-Free Latent Reasoning} We compare against two recent training-free latent reasoning methods: (1)~\textbf{Mixture of Inputs} (MoI; \citealt{su_token_2025}) blends the sampled token's embedding with those from the remaining token distribution; and  (2)~\textbf{Soft Thinking} \citep{zhang_soft_2025} also combines token embeddings, but uses an entropy metric to choose when to end continuous thinking and switch to discrete sampling.

\paragraph{Trained Latent Reasoning} We also compare against two trained latent reasoning methods: (1)~\textbf{COCONUT} \citep{hao_training_2025} trains the whole LLM to use the last hidden state as a token embedding, and applies a curriculum to slowly learn to predict these embeddings instead of reasoning steps; 
and (2)~\textbf{CoLaR} \citep{tan_think_2025},  which uses a non-deterministic latent head to predict compressed embeddings of the next reasoning step, and also has separate SFT and RL stages.  


Finally, we compare against \emph{non-latent} RL, \mbox{\textbf{RL-Trained}}, using GRPO \citep{shao_deepseekmath_2024} on our tasks. This approach represents \textit{upper bound} performance as well as its expected token cost.


We train the full LM for all methods, and use the same hyper-parameters where applicable. Task-specific choices (e.g. COCONUT's max-latent-stage) are defaults or adapted from respective paper advice. More details are in \autoref{sec:hyperparameterdetails}. 

For all trained methods we select the best-performing checkpoint by validation performance. For CoLaR and COCONUT in particular this can produce large variation in token counts between task/model settings, as later checkpoints with greater token savings often collapsed. 

\subsection{NCP Human Evaluation}

Although we optimize next chapter plans via the \textbf{Contrastive Improvement} metric described previously, we are also interested in human judgments of the chapters they induce. We follow the same human evaluation procedure as \citet{gurung_learning_2025}.
We elicit human
preferences on the chapters via pairwise comparisons, along
the dimensions of Plot, Creativity, Development, Language Use,
Characters, and Overall Preference. 

Due to the high cost of human annotations, we compare the following representative sample of methods: Default (untrained Qwen model), \mbox{RL-Trained}, Mixture-of-Inputs, CoLaR, and LiteReason. We generate chapters with Qwen 2.5 7B-Instruct based on the plans generated by each method, and collect 20 pairwise comparisons for every method-combination. We compare model performance by Bradley-Terry relative strengths fit to these comparisons. More details are presented in
\autoref{sec:ncp_human_evals}.

\begin{table*}[t]
\resizebox{\textwidth}{!}{
{\small
\begin{tabular}{@{}lr@{~~}r@{~~}r@{~~}rc@{~~}r@{~~}c@{~~}r@{}}
\toprule
& \multicolumn{4}{c}{\bf Flawed Fictions} & \multicolumn{4}{c}{\bf NCP} \\
 \multicolumn{1}{@{}l}{\bf Method}  & \multicolumn{2}{c}{\textbf{Accuracy (\%)}} & \multicolumn{2}{c}{\textbf{Avg \# Tokens}} & \multicolumn{2}{c}{\textbf{Contra Improve (\%)}} & \multicolumn{2}{c}{\textbf{Avg \# Tokens}} \\
\midrule
Qwen2.5-7B & 57.26 {\tiny $\pm$ 1.58} & &  400.04 {\tiny $\pm$ 6.50} & & 0.067 {\tiny $\pm$ 0.01} & & 831.85 {\tiny $\pm$ 8.26} & \\
MoI & 58.06 {\tiny $\pm$ 1.75} &  (\textcolor{dgreen}{+0.80\%}) & 401.73 {\tiny $\pm$ 6.57} & (\textcolor{red}{+0.42\%}) & 0.045 {\tiny $\pm 0.03$} & (\textcolor{red}{-32.84\%}) & 829.14 {\tiny $\pm 3.13$} & (\textcolor{dgreen}{-2.46\%}) \\
Soft Thinking & 57.42 {\tiny $\pm 1.21$} & (\textcolor{dgreen}{+0.16\%}) & 360.90 {\tiny $\pm 2.96$} & (\textcolor{dgreen}{-9.78\%}) & 0.013 {\tiny $\pm 0.02$} & (\textcolor{red}{-80.75\%}) & 966.73 {\tiny $\pm 4.29$} & (\textcolor{red}{+16.21\%}) \\
\arrayrulecolor{gray} \hline
COCONUT &   50.65 {\tiny $\pm 0.02$}  &(\textcolor{red}{-6.61\%}) &   268.07 {\tiny $\pm 6.68$} &(\textcolor{dgreen}{-32.99\%}) &   0.083       {\tiny $\pm 0.00$}  & (\textcolor{dgreen}{+23.88\%}) &   361.84 {\tiny $\pm 0.92$} &(\textcolor{dgreen}{-56.50\%}) \\
CoLaR & 53.71 {\tiny $\pm 0.01$} & (\textcolor{red}{-3.55\%}) &  226.14 {\tiny $\pm 10.00$} &(\textcolor{dgreen}{-43.47\%}) & 0.118 {\tiny $\pm 0.01$} &(\textcolor{dgreen}{+58.21\%}) & 218.40 {\tiny $\pm 2.54$} &(\textcolor{dgreen}{-73.75\%}) \\
LiteReason &  87.42 {\tiny $\pm$ 0.47} & (\textcolor{dgreen}{+30.16\%}) & 34.01 {\tiny $\pm 0.38$} & (\textcolor{dgreen}{-91.50\%}) & 0.478 {\tiny $\pm 0.01$} &(\textcolor{dgreen}{+613.43\%}) & 193.11 {\tiny $\pm 2.02$} & (\textcolor{dgreen}{-77.50\%}) \\
RL-Trained & 88.71 {\tiny $\pm$ 0.00} & (\textcolor{dgreen}{+31.45\%}) & 114.53 {\tiny $\pm$ 2.01} &(\textcolor{dgreen}{-71.37\%}) & 0.666 {\tiny $\pm 0.01$} &(\textcolor{dgreen}{+894.03\%}) & 721.33 {\tiny $\pm 6.01$} &(\textcolor{dgreen}{-15.97\%}) \\ \bottomrule
\end{tabular}
}}
\caption{Comparing method performance on Flawed Fictions (left) and Next Chapter Prediction (right). The top and bottom block refer to untrained and trained methods respectively. For both tasks LiteReason performs significantly better than all baseline methods, and comes closest to matching non-latent RL performance. Furthermore, LiteReason produces the lowest number of tokens compared to any method, which indicates it has learned a more efficient style of reasoning. Scores are averages across the test set, with {$\pm \text{SEM}$} denoting the standard error of the mean.
Accuracy percent differences are
method - default. All other columns are unbounded values, so we compute percent difference as $\frac{\text{method} - \text{default}}{\text{default}}$.  Contra Improve refers to our Contrastive Improvement reward metric. 
}
\label{tab:combined_method_comparisons}
\end{table*}

\begin{table}[t]
{\small
\begin{center}
\begin{tabular}{@{}ll@{~~}l@{~~}r@{~~}l@{}}
\toprule
\multicolumn{1}{@{}l}{\bf Method}  &\multicolumn{2}{l}{\bf FF Tokens} & \multicolumn{2}{l}{\bf NCP Tokens} \\
\hline
RL & 61.4M & &  108.8M & \\
RL+LiteReason & 29.0M & (\textcolor{dgreen}{-52.8\%}) & 54.9M &  (\textcolor{dgreen}{-49.5\%}) \\
\hline
\end{tabular}
\end{center}
\caption{Comparing the number of discrete tokens generated during RL (GRPO) training with and without LiteReason, across Flawed Fictions (FF) and Next Chapter Prediction (NCP). We find that our method requires about half as many tokens during training. 
}
\label{tab:rl_gen_tokens}
}
\end{table}

\section{Results}
\label{sec:results}

\subsection{What is the Best LiteReason Design?}
  We first perform ablations on the Flawed Fictions and NCP tasks and report results in 
\autoref{tab:name_variant_both_tasks}. 
We benchmark the LiteReason variants against two baselines, both built upon the Qwen2.5-7B-Instruct model: a prompted default model and its RL-trained version. Crucially, neither approach incorporates latent reasoning capabilities.

We find that \textbf{the full LiteReason variant best balances high performance with large token-length reductions}, achieving similar gains to traditional non-latent RL ($\frac{30.16}{31.45}=95.9\%$ and $\frac{613.43}{894.03}=68.6\%$ of the performance increase for Flawed Fictions and NCP, respectively) while generating over three times fewer tokens. LiteReason w/o SFT shows mixed performance: on Flawed Fictions it performs slightly worse than LiteReason but with fewer tokens, but on NCP it performs the highest of our ablations but produces significantly more tokens (622 compared to LiteReason's 193). We hypothesize that the usefulness of periodic SFT will depend on task-specific features like difficulty and dataset size. Different LiteReason hyper-parameters (like swap-ratio and how often the Reasoning Projector is updated) may produce results in between these extremes, allowing practitioners to balance performance and efficiency as desired. 
We also find that LiteReason w/o RP (i.e.,~simply prompting for skipping reasoning steps) increases reasoning length while performing even worse than the RL-Trained baseline. \autoref{tab:name_variant_ff_prompt} shows more comparisons with/without the implicit-thought prompt on different models.

We also briefly explored the usage of our latent thoughts throughout training to see if RL is increasing or decreasing their likelihood. We find that at the beginning of RL training there is a sharp upward trend in latent-thought usage, however  this behaviour often lowers by the end of training. Future work could explore \textit{inducing} latent reasoning by increasing the likelihood of the implicit-token markers or simply changing the prompt.

\subsection{How does LiteReason Compare against other Latent Methods?}


\autoref{tab:combined_method_comparisons} compares the performance of several reasoning methods on Flawed Fictions and NCP tasks, split into non-trained and trained method blocks. Qwen2.5-7B Instruct is considered `default' performance, and the last row shows upper bound performance (Qwen2.5-7B-Instruct trained with RL).

Our results show that  \textbf{LiteReason significantly outperforms existing latent reasoning methods on both tasks.} Relative to the Qwen 2.5 7B-Instruct baseline and all other latent-reasoning methods, LiteReason sees significant performance gains. Accuracy on Flawed Fictions goes up~30.16\% and on NCP Contrastive Improvement goes up \mbox{$>$6$\times$} (\autoref{tab:combined_method_comparisons}). In contrast, the next-best performing latent-reasoning methods only marginally improve performance (0.8\% on Flawed Fictions and 58\% on NCP). We hypothesize this is because our method is most comparable in approach to the RL-Trained baseline, and better explores alternate reasoning paths that the base model would not generate. In contrast, COCONUT and CoLaR methods rely more heavily on the quality of the initial model's reasoning traces, and may overfit on the small datasets.

Unsurprisingly we find that the training-free methods, Mixture-of-Inputs (MoI) and Soft Thinking, both are very close to the untrained model in performance and token length. On the NCP task, Soft Thinking performs marginally better  than the baseline but actually produces slightly more tokens, while MoI performs worse than the baseline with slightly fewer tokens. On the Flawed Fictions task, the differences to the baseline are within confidence intervals. Future research is needed to make the efficiency and performance gains more consistent and pronounced across tasks and reasoning styles for training-free approaches.

\autoref{fig:bradleyterryscores} further examines the quality of generated book chapters on the NCP task through human judgments, comparing the untrained Qwen2.5-7B model against MoI, CoLaR, and a non-latent RL-trained model. We find human judgments (measured as 
relative strengths) largely follow the contrastive improvement trends in \autoref{tab:combined_method_comparisons} with default RL training performing best, followed by LiteReason, and then a notable gap before the remaining methods.

\subsection{Does LiteReason Benefit from RL?}

We also examine more closely  the effect of RL on the relevant methods: non-latent reasoning, CoLaR, and LiteReason \cref{tab:method_rl_comparisons_ff,tab:method_rl_comparisons_ncp} in the Appendix show performance for these methods before and after RL. As mentioned previously, we consider the standard non-latent RL training as an upper bound on performance.

CoLaR is notable in its use of RL \textit{in the latent space}, as opposed to exclusively token-based RL (LiteReason and standard RL training). However, we find CoLaR fails to significantly benefit from RL training, performing only marginally better than the post-SFT model in both performance and efficiency (\cref{tab:method_rl_comparisons_ff,tab:method_rl_comparisons_ncp}). CoLaR before and after RL performs only marginally better than the default model, indicating that they largely compress, not extend, existing model's capabilities. In contrast, \textbf{LiteReason benefits significantly from RL}, moving from near default-model performance to close to the RL-trained upper bound. 

\pgfplotsset{compat=1.18}
\begin{figure}[t]
\begin{tikzpicture}[scale=.60]
    \begin{axis}[
        width  = 0.8*\textwidth,
        height = 7cm,
        major x tick style = transparent,
        ybar=2*\pgflinewidth,
        bar width=8pt,
        ymajorgrids = true,
        ylabel = {Strength},
        ylabel style={yshift=-5pt},
        symbolic x coords={Plot,Character,Creativity,Develop, Language, Overall},
        xtick = data,
        scaled y ticks = false,
        enlarge x limits=0.1,
        ymin=0.15,
        font=\large,
        ymax=2.2,
        legend cell align=left,
        legend style={
                at={(.5,1.05)},
                anchor=north,
                legend columns = -1
        }
    ]
        \addplot[style={bblue,fill=bblue,mark=none}] 
             coordinates {(Plot,0.963) (Character,0.893) (Creativity,0.929) (Develop, 1.053) (Language, 1.081) (Overall, 0.877)};

        \addplot[style={rred,fill=rred,mark=none}] 
             coordinates {(Plot, 0.805) (Character, 0.859) (Creativity, 0.787) (Develop, 0.628) (Language, 0.896) (Overall, 0.832)};

        \addplot[style={ggreen,fill=ggreen,mark=none}] 
             coordinates {(Plot,	0.733) (Character,0.787) (Creativity,0.906) (Develop, 0.752) (Language, 0.879) (Overall, 0.753) };

        \addplot[style={oorange,fill=oorange,mark=none}] 
             coordinates {(Plot,1.095) (Character,1.342) (Creativity,1.066) (Develop, 1.071) (Language, 1.080)  (Overall, 1.233)};

        \addplot[style={ppurple,fill=ppurple,mark=none}] 
             coordinates {(Plot,1.608) (Character,1.235) (Creativity,1.415) (Develop, 1.878) (Language, 1.086)  (Overall, 1.475)};

        \legend{Default, MoI, CoLaR, LiteReason, RL-trained}
    \end{axis}
\end{tikzpicture}
    \caption{Bradley Terry Relative Strength on NCP task.  Default refers to the untrained Qwen2.5-7B model. RL-trained is the upper bound trained with non-latent RL. We find relative strengths largely follow the trend of contrastive improvement in \autoref{tab:combined_method_comparisons}, with  RL-training performing the best, followed by LiteReason, and then a gap and the rest of the methods. }
\label{fig:bradleyterryscores}
\end{figure}
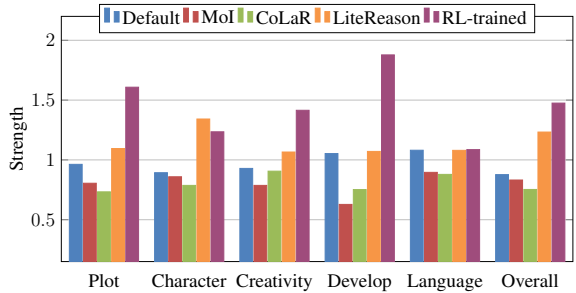

\subsection{Does LiteReason  Improve RL Efficiency?}

\autoref{tab:rl_gen_tokens} shows the number of tokens generated \emph{during RL training} with and without LiteReason.
We find
\textbf{LiteReason requires significantly fewer tokens during RL training}. Although we train for the \emph{same} number of steps and samples, LiteReason uses significantly fewer tokens, 52.8\% fewer on the Flawed Fictions task (29M vs. 61M) and 49.5\% fewer on the NCP task (55M vs. 109M).

In addition to requiring fewer tokens during training, \autoref{tab:combined_method_comparisons} shows that our method also generates substantially fewer tokens during inference: 77\% fewer on the NCP task and 92\% fewer on Flawed Fictions, relative to the base model. Our method is also faster in practice; our wall-clock analysis in \autoref{tab:qwen25_wallclock} shows time-per-example is lower in both single item and large batch settings. 

Taken together, our results in Tables~\ref{tab:combined_method_comparisons}--\ref{tab:qwen25_wallclock} demonstrate that \textbf{LiteReason guides RL to more efficient solutions}. When compared against the RL-Trained model, LiteReason produces traces 73\% smaller on Flawed Fictions and 70\% smaller on NCP. Despite this, our method still achieves 96\% of the performance \textit{gains} on Flawed Fictions and 69\% on NCP relative to non-latent RL training. The performance--cost tradeoff is shown in \autoref{fig:flawed_fiction_pareto} and \autoref{fig:ncp_pareto}, and further discussed in \autoref{sec:performance_cost_curves}. LiteReason thus sits on the same Pareto frontier as the RL-Trained model, while all other methods fall well behind on performance.


\begin{table}[t]
{\small
\begin{center}
\begin{tabular}{@{}llcrr@{}}
\toprule
\textbf{Task} & \textbf{Model} & \textbf{Latent} & \textbf{$b{=}1$} (s) & \textbf{$b{=}\mathrm{all}$} (s) \\
\midrule
FF & Base & \(\times\) & 3.10{\tiny$\pm$0.28} & 0.13{\tiny$\pm$0.02} \\
FF & RL-Trained & \(\times\) & 0.97{\tiny$\pm$0.02} & 0.06{\tiny$\pm$0.00} \\
FF & LiteReason & \(\times\) & 0.83{\tiny$\pm$0.16} & 0.20{\tiny$\pm$0.10} \\
FF & LiteReason & \(\checkmark\) & \textbf{0.31{\tiny$\pm$0.01}} & \textbf{0.03{\tiny$\pm$0.00}} \\
\midrule
NCP & Base & \(\times\) & 8.02{\tiny$\pm$0.35} & 0.43{\tiny$\pm$0.01} \\
NCP & RL-Trained & \(\times\) & 6.46{\tiny$\pm$0.16} & 0.42{\tiny$\pm$0.02} \\
NCP & LiteReason & \(\times\) & 2.21{\tiny$\pm$0.02} & 0.26{\tiny$\pm$0.04} \\
NCP & LiteReason & \(\checkmark\) & \textbf{1.84{\tiny$\pm$0.03}} & \textbf{0.23{\tiny$\pm$0.00}} \\
\bottomrule
\end{tabular}
\end{center}
\caption{\label{tab:qwen25_wallclock}Mean seconds (s) per example on Flawed Fictions (FF) and Next-Chapter-Prediction (NCP) test-sets with the Qwen 2.5 7B models, using vLLM on one H100. LiteReason is tested with and without the latent-reasoning prompt. We report mean and std. dev. seconds/example across 3 sequential repetitions. For $b{=}1$, each repetition runs one example at a time. For $b{=}\mathrm{all}$, each repetition passes the full-dataset for parallel computation. We find that LiteReason produces answers approximately 3x faster than the RL-Trained baseline.}
}
\end{table}

\subsection{
Does LiteReason Retain General Model Capabilities?}


We compare initial model and trained-model performance on GSM-Hard \citep{gao2022pal}, AIME-2025, and MMLU-Redux \citep{gema-etal-2025-done}. The first two tasks test models on computation-heavy and reasoning-heavy math questions respectively. MMLU-Redux is a revised form of MMLU that corrects for errors in the original dataset, and tests for general language understanding across domains.

We evaluate latent-reasoning models in two ways: with standard non-latent discrete decoding and with their respective latent-inference pipelines. This gives us insight into the transferability of learned latent-reasoning. 

Results for the Qwen 2.5-7B based models are shown in \autoref{tab:qwen25_generalization}. We find that LiteReason and our RL-Trained baseline retain the abilities of the base model across tasks. LiteReason models also reason \textit{more concisely}, consistently producing fewer tokens than the base model and RL-Trained variants. Although these efficiency gains are smaller than those in-domain, it indicates some reasoning behaviour may transfer across tasks.

In contrast, the other trained latent-reasoning methods exhibit dramatic drops in performance.  COCONUT, which finetunes the whole model to produce latent tokens, performed significantly worse than the base model in all settings. CoLaR, which has a similar latent-head module to LiteReason, exhibits stable behaviour when trained on Flawed Fictions but significant performance collapse when trained on NCP. We hypothesize that this may be caused by Flawed Fictions following the same \textit{\textbackslash boxed\{\}} format to the other benchmarks but NCP requires long-form answers.

\subsection{
Does LiteReason Work Across Model Size and Family?}
\label{sec:other_models}

We perform experiments with Qwen 3 4B-Instruct-2507, focusing on the Flawed Fictions task and running all trained baselines. Note that these models are much worse at the initial task; their base performance is worse than random at 33\%. We hypothesize that gains from latent reasoning attempts with these models will be smaller, as the initial reasoning quality is very low. We also experiment with the Gemma 3 1B model in our math-task ablation \autoref{sec:math_tasks}.

Results in \autoref{tab:ff_small_models} show similar trends to our previous experiments, but the distinction between LiteReason and RL-Trained is smaller. LiteReason still significantly outperforms the base model in performance and efficiency: the Qwen 3 4B LiteReason model's accuracy increases from $33\%$ to $60$\%, while producing 60\% of the tokens.

Both CoLaR and COCONUT do produce significantly more efficient models, but at the significant cost of worse performance; for example the Qwen 3 4B CoLaR model produces only 336 tokens on average, but also only gets 29\% of questions correct (again, worse than random).

\subsection{
Is LiteReason Compatible with Length-Based Reward Shaping?}

\begin{table}[t]
{\small
\begin{center}
\begin{tabular}{@{}lrr@{}}
\toprule
\textbf{Method} & \textbf{Accuracy (\%)} & \textbf{Avg.\ \# Tokens} \\
\midrule
RL-Trained & 88.71 {\tiny $\pm$ 0.00} & 114.53 {\tiny $\pm$ 2.01} \\
\ \ +LP & 91.94 {\tiny $\pm$ 0.34} & 16.65 {\tiny $\pm$ 4.43} \\
\midrule
LiteReason & 87.42 {\tiny $\pm$ 0.47} & 34.01 {\tiny $\pm$ 0.38} \\
\ \ +LP & \textbf{93.55 {\tiny $\pm$ 0.00}} & \textbf{6.00 {\tiny $\pm$ 0.00}} \\
\bottomrule
\end{tabular}
\end{center}
\caption{Results adding a Length Penalty (LP) to RL-Training on the Flawed Fictions task with Qwen 2.5 7B-Instruct. We find that the penalty improves performance and efficiency, and that combining it with our LiteReason (LR) method \textit{further} improves both.}
\label{tab:qwen25_ff_length_penalty}
}
\end{table}

Another approach to mitigate the costs of RL-training and LM inference is to explicitly reward shorter responses during RL. We experiment with the overlong reward shaping method from DAPO, which showed success in encouraging shorter reasoning traces \citep{yu2025dapoopensourcellmreinforcement}.

Specifically we penalize truncated responses (setting their reward to zero) and add a length-reward $r_{\text{length}} = -0.5\frac{|y|}{\text{max length}}$ on the Qwen 2.5 7B model. We compare adding this penalty to both standard RL training and LiteReason for the Flawed Fiction tasks. We hypothesize that balancing a length penalty reward may be difficult for long-form writing tasks like NCP, but that it may be an appropriate solution to long reasoning traces in a binary prediction task like Flawed Fictions.

Our results (\autoref{tab:qwen25_ff_length_penalty}) show this length penalty term works surprisingly well at balancing performance and efficiency, and that combining it with LiteReason produces not only the most efficient model but the most performant as well. 
We believe this to be an advantage of the RL-centric approach LiteReason takes, as other training techniques can be easily used in conjunction.

\subsection{
Does LiteReason Work on Math Tasks?}
\label{sec:math_tasks}

\begin{table}[t]
{\small
\begin{center}
\begin{tabular}{@{}lrr@{}}
\toprule
\textbf{Method} & \textbf{Accuracy (\%)} & \textbf{Avg. \# Tokens} \\
\midrule
Base & 14.70 {\tiny $\pm$ 0.39} &  932.82 {\tiny $\pm$ 20.28} \\
RL-Trained & \textbf{15.91 {\tiny $\pm$ 0.96}} & 941.13 {\tiny $\pm$ 15.08} \\
LiteReason & 15.76 {\tiny $\pm$ 0.28} & 849.92 {\tiny $\pm$ 19.84} \\
CoLaR & 1.14 {\tiny $\pm$ 0.15} & 978.24 {\tiny $\pm$ 10.88} \\
COCONUT & 1.52 {\tiny $\pm$ 0.00} & 25.12 {\tiny $\pm$ 0.02} \\
\bottomrule
\end{tabular}
\end{center}
\vspace{-8pt}
\caption{Results with Gemma-3-1B-IT models on GSM-Hard. The RL-Trained model only performs 6\% better, and LiteReason performs largely the same as the base model while producing 10\% fewer tokens. CoLaR and COCONUT models both collapse to worse performance, possibly due to poor initial reasoning quality. These results mirror trends from the narrative tasks, highlighting LiteReason as a promising latent-reasoning method beyond narrative tasks.}
\label{tab:gemma1b_gsm_hard}
}
\vspace{-10pt}
\end{table}

Although LiteReason is designed around narrative task constraints, we evaluate its performance on a more traditional testbed for latent-reasoning: GSM-Hard's math benchmark \citep{gao2022pal}.

We use Gemma 3 1B-IT as the basis for our experiments, which has a low initial accuracy of $14.7\%$ while generating $932.82$ tokens on average. We base all experiments on Flawed Fictions hyper-parameters, with the exception of RL-epochs (lowered to 5) and group-size (lowered to 8).

\autoref{tab:gemma1b_gsm_hard} shows similar trends to our narrative tasks, but with smaller differences from the base model for LiteReason and RL-Training, and dramatic performance collapses for CoLaR and COCONUT models. RL-training (our upper bound) only increases performance to $15.91\%$, with the slightly higher $941.13$ tokens. 

LiteReason slightly improves performance from the base model, achieving $15.76\%$ with just $849.92$ tokens (a 10\% token reduction). In contrast, CoLaR and COCONUT reduce to $\approx1\%$ performance, though COCONUT does significantly reduce tokens to $25.12$.

Thus LiteReason shows promise on other difficult tasks more commonly used in latent-reasoning research, and indicates greater robustness to poor initial reasoning abilities.

\section{Conclusion}
\label{conclusion}

We presented LiteReason, a latent-reasoning method designed for narrative domains that integrates well with RL training. LiteReason trains a lightweight Reasoning Projector to produce informative latent tokens and teaches an LLM to interleave these latent thoughts with discrete sampling.

We evaluate our method against several training-free and trained latent reasoning baselines  on two challenging tasks (Flawed Fictions and Next Chapter Prediction) that require long-context understanding and diverse reasoning. We find that LiteReason achieves the closest performance to traditional RL training while requiring less than half the generated tokens during training and shrinking the reasoning generated at inference time by 70\%. We also show our trained model maintains other capabilities better than existing latent reasoning methods. We perform ablations on other model families and the GSM-Hard task, and show LiteReason's consistent performance-efficiency balancing behaviour.

To our knowledge, we are the first to apply continuous latent reasoning to narrative tasks and the first to interleave latent reasoning with discrete token sampling during RL, letting the model freely decide when to use latent-tokens during training. In the future, we hope to explore alternate Reasoning Projector designs, such as recursively updating latent thoughts or injecting latent thoughts at locations other than the token embedding stage. 

\section*{Acknowledgments}

This work was supported by the Edinburgh International Data Facility (EIDF) and the Data-Driven Innovation Programme at the University of Edinburgh. Access to EIDF was facilitated through the University of Edinburgh’s Generative AI Laboratory GAIL Fellow scheme. ESW acknowledges support from the CIFAR Learning in Machines and Brains Programme.

\bibliography{tacl26latent}
\bibliographystyle{acl_natbib}

\clearpage

\appendix

\section{Extending VR-CLI with Contrastive Reward}
\label{sec:extending_vrcli}

In this section we build on the VR-CLI objective proposed in \cite{gurung_learning_2025} to add a contrastive term, which we use as our primary automated metric for our NCP results. We use the formalisms introduced in the original paper: at a chapter index~$i$, a \emph{reasoning model}
$\pi_{\theta}^{\mathcal{R}}$ reasons over Story Information~$SI_i$ and
predicts a plan $\hat{p} \leftarrow \pi^{\mathcal{R}}_\theta (SI_i)$ for the next chapter. $SI_i$~contains a
global story sketch, previous chapters plot summary and character sheets, the preceding chapter, and
a high-level synopsis of the next chapter. 

As discussed, we find significant repetitions in the traces from the original work's objective. These kinds of reasoning traces would not serve as a useful test-bed for latent reasoning, as they could be easily compressed (and writing advice could be folded into the prompt). We propose a simple addition to the VR-CLI formulation to produce significantly more diverse reasoning traces after training.  We call our addition a \textbf{contrastive term}, which consists of subtracting a (weighted) improvement term for another chapter from the original improvement. 

For each group we get the likelihood improvement, given the generated plan, for a randomly selected chapter from a different book. We subtract this term to penalize plans that give general advice that could be applicable to any chapter. Because each element within a group receives the same random chapter, we do not find destabilizing effects.

\noindent\begin{align*}
    R_\text{old} &= I(SI_i, c_{i+1}, \hat{p}) \\
    R_\text{contr} &= I(SI_i, c_{i+1}, \hat{p}) - \gamma I(SI_i, c^{r}, \hat{p})
\end{align*}

\autoref{tab:diversity_contrastive} shows training with this approach produces more diverse plans; our NCP results use this contrastive-improvement as the training and evaluation objective. \autoref{tab:compare_plans_contrastive} shows example plans produced from models trained with and without the contrastive reward. Future work could further explore this approach, modifying the weight (we always use $\gamma=0.5$) and changing the chapter-selection process. For example, down-weighting the likelihood of other chapters in the same book may also have a positive effect by requiring the model to generate plans specific to the current events, but may also discourage the model from referencing true facts about shared characters.

\section{Effect of Prompting and RL on Method Performance}

Table~\ref{tab:name_variant_ff_prompt} shows that removing the implicit-thought prompt from LiteReason variants that use the Reasoning Projector yields slightly higher accuracy at the cost of longer reasoning traces, suggesting the prompt can be used to tune the performance--efficiency tradeoff at test time. Tables~\ref{tab:method_rl_comparisons_ff} and~\ref{tab:method_rl_comparisons_ncp} compare method performance before and after RL on Flawed Fictions and NCP respectively, confirming that LiteReason benefits substantially from RL training while CoLaR does not.

\begin{table*}[t]
{\small
\begin{center}
{%
\begin{tabular}{@{}llcrrrrrr}
\toprule
 \multicolumn{3}{r@{}}{\textbf{Evaluation task $\rightarrow$}} & \multicolumn{2}{c}{\textbf{GSM-Hard}} & \multicolumn{2}{c}{\textbf{AIME25}} & \multicolumn{2}{c}{\textbf{MMLU-Redux}} \\
\cmidrule(lr){4-5} \cmidrule(lr){6-7} \cmidrule(lr){8-9}
\textbf{Task} & \textbf{Method} & \textbf{Latent} & \textbf{Acc.} & \textbf{Tok.} & \textbf{Acc.} & \textbf{Tok.} & \textbf{Acc.} & \textbf{Tok.} \\
\midrule
\multicolumn2{@{}l}{\textit{No finetuning}} & $\times$ & 27.3 {\tiny $\pm$ 0.2} & 371.4 {\tiny $\pm$ 2.0} & 11.3 {\tiny $\pm$ 2.0} & 891.1 {\tiny $\pm$ 10.1} & 78.7 {\tiny $\pm$ 0.1} & 317.0 {\tiny $\pm$ 1.3} \\
\midrule
FF & RL-Trained & $\times$ & 27.2 {\tiny $\pm$ 0.1} & 359.1 {\tiny $\pm$ 0.8} & 10.0 {\tiny $\pm$ 1.8} & 857.3 {\tiny $\pm$ 20.8} & 79.0 {\tiny $\pm$ 0.2} & 300.5 {\tiny $\pm$ 0.5} \\
 & LiteReason & $\times$ & 27.3 {\tiny $\pm$ 0.1} & 349.7 {\tiny $\pm$ 2.1} & 10.7 {\tiny $\pm$ 1.2} & 834.7 {\tiny $\pm$ 9.0} & 78.8 {\tiny $\pm$ 0.0} & 283.4 {\tiny $\pm$ 0.5} \\
 & LiteReason & \(\checkmark\) & 26.9 {\tiny $\pm$ 0.1} & 316.3 {\tiny $\pm$ 2.8} & 4.7 {\tiny $\pm$ 1.7} & 857.4 {\tiny $\pm$ 20.4} & 78.6 {\tiny $\pm$ 0.2} & 273.4 {\tiny $\pm$ 1.0} \\
 & CoLaR & $\times$ & 27.1 {\tiny $\pm$ 0.2} & 372.2 {\tiny $\pm$ 1.3} & 8.0 {\tiny $\pm$ 2.5} & 876.1 {\tiny $\pm$ 9.1} & 78.9 {\tiny $\pm$ 0.2} & 328.4 {\tiny $\pm$ 1.1} \\
 & CoLaR & \(\checkmark\) & 26.9 {\tiny $\pm$ 0.2} & 338.8 {\tiny $\pm$ 1.5} & 8.0 {\tiny $\pm$ 1.7} & 865.8 {\tiny $\pm$ 29.1} & 78.2 {\tiny $\pm$ 0.1} & 208.6 {\tiny $\pm$ 0.5} \\
 & COCONUT & $\times$ & 7.9 {\tiny $\pm$ 0.6} & 461.1 {\tiny $\pm$ 20.1} & 0.0 {\tiny $\pm$ 0.0} & 489.4 {\tiny $\pm$ 82.9} & 29.9 {\tiny $\pm$ 1.9} & 404.6 {\tiny $\pm$ 33.8} \\
 & COCONUT & \(\checkmark\) & 12.2 {\tiny $\pm$ 0.2} & 393.4 {\tiny $\pm$ 9.1} & 0.7 {\tiny $\pm$ 0.7} & 660.0 {\tiny $\pm$ 25.1} & 35.7 {\tiny $\pm$ 0.4} & 344.2 {\tiny $\pm$ 2.4} \\
\midrule
NCP & RL-Trained & $\times$ & 27.3 {\tiny $\pm$ 0.3} & 365.0 {\tiny $\pm$ 1.8} & 10.0 {\tiny $\pm$ 2.8} & 912.7 {\tiny $\pm$ 16.4} & 78.8 {\tiny $\pm$ 0.2} & 297.9 {\tiny $\pm$ 1.0} \\
 & LiteReason & $\times$ & 26.5 {\tiny $\pm$ 0.1} & 334.4 {\tiny $\pm$ 1.2} & 4.0 {\tiny $\pm$ 2.4} & 737.2 {\tiny $\pm$ 10.7} & 78.2 {\tiny $\pm$ 0.1} & 250.0 {\tiny $\pm$ 0.4} \\
 & LiteReason & \(\checkmark\) & 26.5 {\tiny $\pm$ 0.1} & 301.1 {\tiny $\pm$ 1.4} & 7.3 {\tiny $\pm$ 1.9} & 617.1 {\tiny $\pm$ 10.0} & 76.7 {\tiny $\pm$ 0.1} & 186.9 {\tiny $\pm$ 0.5} \\
 & CoLaR & $\times$ & 5.9 {\tiny $\pm$ 0.6} & 716.0 {\tiny $\pm$ 28.1} & 0.7 {\tiny $\pm$ 0.7} & 655.0 {\tiny $\pm$ 53.9} & 26.9 {\tiny $\pm$ 0.7} & 424.0 {\tiny $\pm$ 28.0} \\
 & CoLaR & \(\checkmark\) & 1.3 {\tiny $\pm$ 0.1} & 674.9 {\tiny $\pm$ 12.1} & 0.7 {\tiny $\pm$ 0.7} & 516.9 {\tiny $\pm$ 77.9} & 29.4 {\tiny $\pm$ 0.5} & 660.3 {\tiny $\pm$ 6.3} \\
 & COCONUT & $\times$ & 19.4 {\tiny $\pm$ 0.5} & 349.1 {\tiny $\pm$ 6.8} & 2.0 {\tiny $\pm$ 0.8} & 783.5 {\tiny $\pm$ 53.2} & 43.1 {\tiny $\pm$ 0.4} & 274.8 {\tiny $\pm$ 4.7} \\
 & COCONUT & \(\checkmark\) & 15.2 {\tiny $\pm$ 0.1} & 422.2 {\tiny $\pm$ 7.0} & 0.0 {\tiny $\pm$ 0.0} & 829.6 {\tiny $\pm$ 30.3} & 38.9 {\tiny $\pm$ 0.2} & 376.5 {\tiny $\pm$ 2.0} \\
\bottomrule
\end{tabular}
}
\end{center}
\vspace{-8pt}
\caption{Cross-task generalization test with/without latent reasoning, for Qwen2.5-7B models. 'Latent' means the appropriate latent-inference pipeline was used (i.e. prompting for implicit tokens for LiteReason, using the original inference pipelines for CoLaR and COCONUT), otherwise the models were passed to the standard vLLM-based sampling pipeline. $\pm$ refers to run-level SEM.
We find little performance degradation from RL-based training methods, but non-LiteReason latent models exhibit much worse performance. In general latent-reasoning exhibits mixed cross-task generalization, and cannot be consistently relied upon to improve performance or efficiency. Though the effects are small, we hypothesize that the generalization capability of LiteReason's latent reasoning may correlate with the reasoning style that it is trying to mimic. For example, when trained on longer NCP data we see some small improvements on the similarly long-form reasoning of AIME25 (relative to the non-latent mode), but when trained on the more structured and succinct FF we see performance decrease. Token savings follow a similar trend, with the largest decrease in token usage occurring with the NCP-trained model with latent reasoning.}
\label{tab:qwen25_generalization}
}
\end{table*}

\begin{table}[t]
{\small
\begin{center}
\begin{tabular}{@{}lrr}
\toprule
\textbf{Method} & \textbf{Acc. (\%)} & \textbf{\# Tokens} \\
\midrule
Base & 33.23 {\tiny $\pm$ 1.21} & 1709.86 {\tiny $\pm$ 17.05} \\
RL-Trained & \textbf{64.84 {\tiny $\pm$ 1.64}} & 1078.57 {\tiny $\pm$ 7.28} \\
LiteReason & 57.42 {\tiny $\pm$ 2.42} & 995.38 {\tiny $\pm$ 17.09} \\
CoLaR & 29.35 {\tiny $\pm$ 0.60} & 336.49 {\tiny $\pm$ 22.46} \\
COCONUT & 28.55 {\tiny $\pm$ 1.65} & \textbf{20.64 {\tiny $\pm$ 0.44}} \\
\bottomrule
\end{tabular}
\end{center}
\vspace{-8pt}
\caption{\label{tab:ff_small_models}Results on Flawed Fictions with Qwen3-4B-Instruct-2507 based models. Similar to our experiments with the larger Qwen2.5-7B model, we find that LiteReason performs the closest to standard RL, while CoLaR and COCONUT fail to improve much over the base model despite their efficiency gains.}
}
\vspace{-15pt}
\end{table}

\begin{table*}[t]
{\small
\begin{center}
\begin{tabular}{lclrrr}
\toprule
\multicolumn{1}{l}{\bf Method} & \multicolumn{1}{c}{\bf Implicit Prompt} & \multicolumn{2}{c}{\bf Accuracy (\%)} & \multicolumn{2}{c}{\bf Avg \# Tokens} \\
\midrule
Qwen2.5-7B-Instruct & $\times$ & 57.26 {\tiny $\pm 1.58$} & & 400.04 {\tiny $\pm 6.50$} &  \\
Qwen2.5-7B-Instruct & \(\checkmark\) &  50.65 {\tiny $\pm 1.85$} & (\textcolor{red}{-6.61\%}) & 499.21 {\tiny $\pm 9.20$} &  (\textcolor{red}{+24.79\%}) \\
\arrayrulecolor{gray}\hline
RL-Trained & $\times$ & 88.71 {\tiny $\pm 0.00$} & (\textcolor{dgreen}{+31.45\%}) & 114.53 {\tiny $\pm 2.01$} & (\textcolor{dgreen}{-72.46\%}) \\ 
RL-Trained & \(\checkmark\) & 90.32 {\tiny $\pm 0.00$} & (\textcolor{dgreen}{+33.06\%}) & 144.88 {\tiny $\pm 2.03$} &  (\textcolor{dgreen}{-63.78\%}) \\ 
\arrayrulecolor{gray}\hline
LiteReason & \(\checkmark\) & 87.42 {\tiny $\pm 0.47$} & (\textcolor{dgreen}{+30.16\%}) &  34.01 {\tiny $\pm 0.38$} & (\textcolor{dgreen}{-91.50\%}) \\
LiteReason & $\times$ & 89.68 {\tiny $\pm 0.36$} & (\textcolor{dgreen}{+32.42\%}) & 105.80 {\tiny $\pm 8.46$} & (\textcolor{dgreen}{-73.55\%}) \\
~~~w/o Periodic SFT & \(\checkmark\) & 85.48 {\tiny $\pm 0.00$} & (\textcolor{dgreen}{+28.22\%}) & 8.26 {\tiny $\pm 0.17$} & (\textcolor{dgreen}{-97.94\%}) \\
~~~w/o Periodic SFT & \(\times\) &  87.26 {\tiny $\pm 0.16$} &  (\textcolor{dgreen}{+30.00\%}) & 30.57 {\tiny $\pm 1.98$} &  (\textcolor{dgreen}{-92.36\%}) \\
~~~w/o Reasoning Projector & \(\checkmark\) & 87.58 {\tiny $\pm 0.76$} &  (\textcolor{dgreen}{+30.32\%}) & 261.69 {\tiny $\pm 3.72$} & (\textcolor{dgreen}{-34.58\%}) \\
~~~w/o Reasoning Projector & \(\times\) &  86.61 {\tiny $\pm 0.76$} & (\textcolor{dgreen}{+29.35\%}) & 258.92 {\tiny $\pm 3.35$} & (\textcolor{dgreen}{-35.28\%}) \\
\bottomrule
\end{tabular}
\end{center}
\caption{Comparing LiteReason variants on the Flawed Fictions task, with and without the implicit-thought prompt. Flawed Fictions accuracy percent differences are $\text{new} - \text{old}$, while other metrics percent differences are $\frac{\text{new} - \text{old}}{\text{old}}$. We find slight increases in performance for LiteReason variants that use the Reasoning Projector when we switch to the non-implicit-thought prompt, at the cost of much longer reasoning traces. Future work could leverage this prompting ability or sampling to encourage desired performance-compute tradeoff at test-time.
}
\label{tab:name_variant_ff_prompt}
}
\end{table*}

\begin{table*}[t]
{\small
\begin{center}
\begin{tabular}{@{}ll@{~~}rr@{~~}rr@{~~}}
\toprule
 & \multicolumn{4}{c}{\bf Flawed Fictions} \\
\textbf{Method} & \multicolumn{2}{c}{\bf Accuracy (\%)} & \multicolumn{2}{c}{\bf Avg \# Tokens} \\
\midrule
Qwen2.5-7B-Instruct & 57.26 {\tiny $\pm  1.58$}  & & 400.04 {\tiny $\pm  6.50$}  \\
RL-Trained & 88.71 {\tiny $\pm  0.00$} & (\textcolor{dgreen}{+31.45\%}) & 114.53 {\tiny $\pm  2.01$} &(\textcolor{dgreen}{-71.37\%}) \\
\arrayrulecolor{gray}\hline
LiteReason pre-RL & 52.10 {\tiny $\pm  2.50$} & (\textcolor{red}{-5.16\%}) & 567.63 {\tiny $\pm  13.86$} &(\textcolor{red}{+41.89\%}) \\
LiteReason &  87.42 {\tiny $\pm  0.47$} & (\textcolor{dgreen}{+30.16\%}) & 34.01 {\tiny $\pm 0.38$} & (\textcolor{dgreen}{-91.50\%}) \\
\arrayrulecolor{gray}\hline
CoLaR & 50.89 {\tiny $\pm  0.02$} &(\textcolor{red}{-6.37\%}) &  226.88  {\tiny $\pm 6.33$} &(\textcolor{dgreen}{-43.29\%}) \\
CoLaR-PostRL & 53.71 {\tiny $\pm 0.01$} & (\textcolor{red}{-3.55\%}) &  226.14 {\tiny $\pm 10.00$} &(\textcolor{dgreen}{-43.47\%})  \\ \bottomrule
\end{tabular}
\end{center}
\caption{Comparing performance on Flawed Fictions before and after RL. LiteReason pre-RL refers to using the implicit-thought prompt and the pretrained Reasoning Projector, immediately after its SFT initialization. We find that for both the base model and LiteReason, RL improves performance and reduces tokens. However, CoLaR-PostRL achieves essentially the same performance and length for Flawed Fictions and worse performance and length for NCP, indicating that it is not able to use RL effectively. Flawed Fictions accuracy percent $\pm$ are $\text{new} - \text{old}$, while other metrics percent $\pm$ are $\frac{\text{new} - \text{old}}{\text{old}}$.
}
\label{tab:method_rl_comparisons_ff}
}
\end{table*}

\begin{table*}[t]
{\small
\begin{center}
\begin{tabular}{@{}ll@{~~}rr@{~~}rr@{~~}}
\toprule
 & \multicolumn{4}{c}{\bf NCP} \\
\textbf{Method} & \multicolumn{2}{c}{\bf Contra Improve} & \multicolumn{2}{c}{\bf Avg \# Tokens} \\
\midrule
Qwen2.5-7B-Instruct & 0.067 {\tiny $\pm 0.01$} & & 831.85 {\tiny $\pm 3.13$}   \\
RL-Trained & 0.666 {\tiny $\pm 0.01$} &(\textcolor{dgreen}{+894.03\%}) & 721.33 {\tiny $\pm 6.01$} &(\textcolor{dgreen}{-13.29\%}) \\
\arrayrulecolor{gray}\hline
LiteReason pre-RL & 0.041 {\tiny $\pm 0.03$} & (\textcolor{red}{-38.80\%}) & 946.47 {\tiny $\pm 31.36$} &  (\textcolor{red}{+13.78\%}) \\
LiteReason &  0.478 {\tiny $\pm 0.01$} &(\textcolor{dgreen}{+613.43\%}) & 193.11 {\tiny $\pm 2.02$} & (\textcolor{dgreen}{-76.79\%}) \\
\arrayrulecolor{gray}\hline
CoLaR & 0.118 {\tiny $\pm 0.01$} &(\textcolor{dgreen}{+58.21\%}) & 218.40 {\tiny $\pm 2.54$} &(\textcolor{dgreen}{-73.75\%}) \\
CoLaR-PostRL & 0.118 {\tiny $\pm 0.01$} &(\textcolor{dgreen}{+58.21\%}) & 218.40 {\tiny $\pm 2.54$} &(\textcolor{dgreen}{-73.75\%}) \\ \bottomrule
\end{tabular}
\end{center}
\caption{Comparing performance on the Next Chapter Prediction task before and after RL. LiteReason pre-RL refers to using the implicit-thought prompt and the pretrained Reasoning Projector, immediately after its SFT initialization. We find that for both the base model and LiteReason, RL improves performance and reduces tokens. However, CoLaR-PostRL achieves essentially the same performance and length for Flawed Fictions and never improves on the SFT model for NCP (and thus the `best' model is the same), indicating that it is not able to use RL effectively. Contrastive Improvement and Average Tokens comparisons inside ($\pm$) are $\frac{\text{new} - \text{old}}{\text{old}}$ 
}
\label{tab:method_rl_comparisons_ncp}
}
\end{table*}
 
\section{Plotting Performance-Cost Curves}
\label{sec:performance_cost_curves}

\begin{figure}[t]
\begin{center}
\begin{tikzpicture}[scale=.52]
\begin{axis}[
    width=14cm,
    height=10cm,
    xlabel={Contrastive Improvement},
    ylabel={Generated Tokens},
    xlabel style={font=\Large},
    ylabel style={font=\Large},
    xmin=0, xmax=0.75,
    ymin=150, ymax=1050,
    grid=none,
    tick label style={font=\large},
    legend style={
        at={(.72,0.30)}, 
        anchor=west,
        font=\large,
        cells={anchor=west},
        draw=black,
        fill=white,
        rounded corners,
        row sep=2.5pt,
    },
    legend cell align={left},
]

\addplot[
    only marks,
    mark=*,
    mark size=6pt,
    color=blue!70!black,
    mark options={solid, fill=blue!70!black},
] coordinates {(0.067, 831.85)};
\addlegendentry{Qwen2.5-7B}

\addplot[
    only marks,
    mark=x,
    mark size=8pt,
    line width=2pt,
    color=orange!80!red,
] coordinates {(0.045, 829.14)};
\addlegendentry{MoI}

\addplot[
    only marks,
    mark=square*,
    mark size=5pt,
    color=green!60!black,
    mark options={solid, fill=green!60!black},
] coordinates {(0.013, 966.73)};
\addlegendentry{Soft Thinking}

\addplot[
    only marks,
    mark=+,
    mark size=7pt,
    line width=2.5pt,
    color=red!70!black,
] coordinates {(0.083, 361.84)};
\addlegendentry{COCONUT}

\addplot[
    only marks,
    mark=diamond*,
    mark size=8pt,
    color=purple!70!blue,
    mark options={solid, fill=purple!70!blue},
] coordinates {(0.118, 218.4)};
\addlegendentry{CoLaR}

\addplot[
    only marks,
    mark=diamond*,
    mark size=8pt,
    color=brown!70!red,
    mark options={solid, fill=brown!70!red},
] coordinates {(0.478, 193.11)};
\addlegendentry{LiteReason}

\addplot[
    only marks,
    mark=triangle*,
    mark size=8pt,
    color=magenta!70!red,
    mark options={solid, fill=magenta!70!red},
] coordinates {(0.666, 721.33)};
\addlegendentry{RL-Trained}

\end{axis}
\end{tikzpicture}

\end{center}
\caption{Generated tokens vs Contrastive Improvement for the Next Chapter Prediction task, by method, for Qwen 2.5 7B models. Better methods are to the right (higher Contrastive Improvement) and lower (fewer tokens). We find LiteReason performs significantly more efficiently than the standard RL-Trained model, with only slightly worse performance. Aside from RL-Trained, there is a significant gap in both Contrastive Improvement and efficiency between LiteReason and the next best method. Thus, we claim our model sits on the same Pareto frontier as the RL-Trained baseline.}
\label{fig:ncp_pareto}
\end{figure}
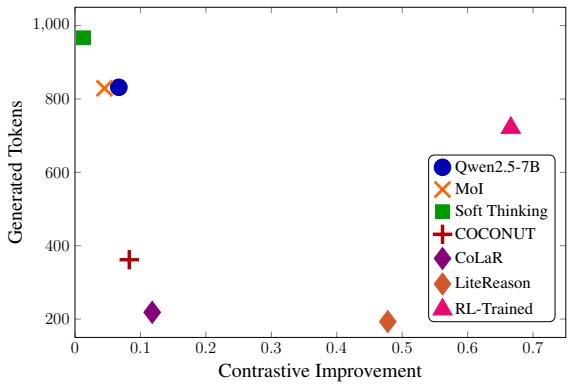

We plot models by their performance (i.e.,~reward on the given task) and cost (measured by the generated token length). Our goal is to show that all methods lie somewhere in this tradeoff space, and that LiteReason pushes the Pareto front by achieving comparable performance to non-latent RL methods while being significantly less costly. \autoref{fig:ncp_pareto} shows this tradeoff curve for the NCP task, and \autoref{fig:flawed_fiction_pareto} for Flawed Fictions.

\begin{figure}[t]
\begin{center}
\begin{tikzpicture}[scale=.52]
\begin{axis}[
    width=14cm,
    height=10cm,
    xlabel={Accuracy},
    ylabel={Generated Tokens},
    xlabel style={font=\Large},
    ylabel style={font=\Large},
    xmin=45, xmax=95,
    ymin=0, ymax=450,
    grid=none,
    tick label style={font=\large},
    legend style={
        at={(.7,0.72)}, 
        anchor=west,
        font=\large,
        cells={anchor=west},
        draw=black,
        fill=white,
        rounded corners,
        row sep=2.5pt,
    },
    legend cell align={left},
]

\addplot[
    only marks,
    mark=*,
    mark size=6pt,
    color=blue!70!black,
    mark options={solid, fill=blue!70!black},
] coordinates {(57.26, 400.04)};
\addlegendentry{Qwen2.5-7B}

\addplot[
    only marks,
    mark=x,
    mark size=8pt,
    line width=2pt,
    color=orange!80!red,
] coordinates {(58.06, 401.73)};
\addlegendentry{MoI}

\addplot[
    only marks,
    mark=square*,
    mark size=5pt,
    color=green!60!black,
    mark options={solid, fill=green!60!black},
] coordinates {(57.42, 360.9)};
\addlegendentry{Soft Thinking}

\addplot[
    only marks,
    mark=+,
    mark size=7pt,
    line width=2.5pt,
    color=red!70!black,
] coordinates {(50.65, 268.07)};
\addlegendentry{COCONUT}

\addplot[
    only marks,
    mark=diamond*,
    mark size=8pt,
    color=purple!70!blue,
    mark options={solid, fill=purple!70!blue},
] coordinates {(53.71, 226.14)};
\addlegendentry{CoLAR}

\addplot[
    only marks,
    mark=diamond*,
    mark size=8pt,
    color=brown!70!red,
    mark options={solid, fill=brown!70!red},
] coordinates {(87.42, 34.01)};
\addlegendentry{LiteReason}

\addplot[
    only marks,
    mark=triangle*,
    mark size=7pt,
    color=magenta!70!red,
    mark options={solid, fill=magenta!70!red},
] coordinates {(88.71, 114.53)};
\addlegendentry{RL-Trained}

\end{axis}
\end{tikzpicture}
\end{center}
\caption{Generated tokens vs Accuracy for the Flawed Fictions benchmark, by method, for Qwen 2.5 7B models. Better methods are to the right (more accurate) and lower (fewer tokens). We find LiteReason performs significantly more efficiently than the standard RL-Trained model, with only slightly worse performance. 
Aside from RL-Trained, there is a significant gap in both accuracy (about 20\%) and efficiency (about 100 tokens) between LiteReason and the next best method. Thus, we claim our model sits on the same Pareto frontier as the RL-Trained baseline.
}
\label{fig:flawed_fiction_pareto}
\end{figure}
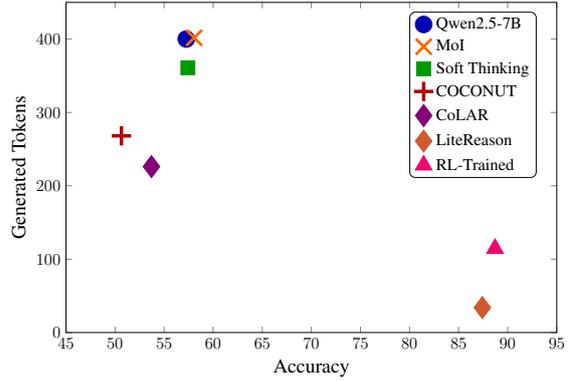

\clearpage
\clearpage

\section{Training and Hyper-parameter Details}
\label{sec:hyperparameterdetails}


We report justifications for training/inference hyper-parameters, and details concerning how we report token counts, computational cost, and statistical significance. We also provide prompts with and without implicit-prompt instructions: Flawed Fictions in \autoref{tab:flawed_fiction_prompts} and Next Chapter Prediction in \autoref{tab:ncp_prompts}.


Hyper-parameters were chosen from prior work (e.g. using RL hyper-parameters from \citet{gurung_learning_2025}) and initial experiments on NCP and FF by validation performance. Due to their long-context, training on these tasks is very computationally expensive; we hope future work gives more guidance on hyper-parameter selection. 

For \textbf{VR-CLI}, contrastive weight $\gamma{=}0.5$. Details about this term are presented in \autoref{sec:extending_vrcli}.

{For the \textbf{length-penalty} discussed in our length-based reward shaping ablation, we penalize truncated responses (setting their reward to zero) and add a length-reward $r_{\text{length}} = -0.5\frac{|y|}{\text{max length}}$. Although we did not run a sensitivity analysis, we anticipate the efficacy of this term to be dependent on the reward distribution of the model on one's dataset. Because we provide negative reward for every additional token (instead of providing a heuristically-set `buffer' of unpenalized tokens), if the model's reward is uniform across an advantage group the only reward signal is length. As a result, if the model fails to find any correct answers for a question it may be pushed to shorter and shorter \textit{incorrect} answers, collapsing training. We did not find this behaviour in our experiments, potentially because the binary nature of the Flawed-Fictions task increases the likelihood that any sample in the group found the correct answer.}

\subsection*{Trained Methods}

For \textbf{LiteReason}, learning rates match CoLaR (1e-4 SFT, 5e-7 RL),
with max generation length 2048, token-replacement ratio $t_r{=}0.2$,
and sentence-replacement ratio $s_r{=}(10\%, 25\%)$.
FF uses 1 epoch for SFT, 30 epochs for RL, batch size 48, and 16 samples per group;
NCP uses 2 epochs for SFT, 20 epochs for RL, batch size 64, and 8 samples per group. The NCP training hyper-parameters are based on \citep{gurung_learning_2025}, and Flawed Fictions hyper-parameters were adjusted to train for longer and with larger groups to account for the smaller dataset size.

LiteReason only uses a fixed number of latent steps for each implicit-thought segment, not a globally fixed reasoning length. During SFT, the number of latent steps is derived from the length of the replaced reasoning span using the token-replacement ratio $t_r$; during inference, the model emits an implicit-thought tag with an integer budget and then returns to discrete decoding after that many Reasoning Projector steps. This design avoids learning a separate halting policy, keeping computation predictable and making latent mode easy to combine with token-space RL. {The main tradeoff is that small budgets may under-represent information-dense reasoning spans, while larger budgets reduce the efficiency benefit and increase the cost of backpropagating through latent rollouts. With $t_r=0.2$, roughly one latent token is used for every five original reasoning tokens. We chose this compression ratio from observation of the narrative-focused reasoning traces, which tend to use more full, descriptive sentences relative to traditional latent reasoning tasks like math. However, we also cap the number of latent reasoning tokens per replaced-sentence at 5, to prevent long, likely verbose, sentences from dominating training and teaching inefficient reasoning.}

For \textbf{CoLaR}, we use a learning rate of 1e-4 (SFT) and 5e-7 (RL),
a compression factor of 5, max generation length of 2048, and 64 max latents.
FF uses 5 epochs for SFT, 30 epochs for RL, batch size 48, and 16 samples per group;
NCP uses 2 epochs for SFT, 20 epochs for RL, batch size 64, and 8 samples per group. These are set to match the LiteReason decisions to ensure models are trained for the same steps and shown the same number of samples. The max compression factor is based on the CoLaR source code default from \citet{tan_think_2025}.

For \textbf{COCONUT}, both use a learning rate of 5e-5, 14 epochs,
1 continuous thought per reasoning step, 2 epochs per stage, and 10 latent stages;
batch sizes are 48 (FF) and 64 (NCP). These hyper-parameters were chosen to best match the amount of training signal from RL-based methods; for example this setup runs for 33 epochs which gives a total epoch count more comparable to the SFT+RL training found in CoLaR and LiteReason. Continuous thought was set to 1 as \citet{hao_training_2025} uses this value for logical reasoning, the most similar task. We initially attempted a learning rate of 1e-4 to match the other SFT stages (and COCONUT defaults), but found that performance collapsed and gave validation performance of 0. The original work trains on significantly smaller models (GPT-2) and with much larger datasets, which may have counter-acted this behaviour.

In contrast to the original work by \citet{hao_training_2025}, we select the best checkpoint by validation performance across all epochs. This better aligns with our goal to train models that sit on the performance-efficiency tradeoff, and prevents us from unfairly selecting low-accuracy checkpoints with high efficiency but near zero performance.



\subsection*{Training-Free}

For \textbf{Mixture-of-Inputs} We evaluated performance on the validation set while sweeping the $\beta$ hyper-parameter $[0.25, 0.5, 1.0, 2.0, 4.0]$ based on the advice presented in \citet{zhuang_text_2025}. We selected the best performing $\beta$ for test-set evaluation ($0.5$ for Flawed Fictions, $4.0$ for NCP).

For training-free methods (\textbf{Soft-Thinking and Mixture-of-Inputs}) we perform a small hyper-parameter sweep across temperature (0.6, 0.7), top-p (0.8, 0.95), and top-k (20, 30) on the validation set, following previous work \citep{su_token_2025,zhang_soft_2025}. For Soft-Thinking we also test gumbel and dirichlet (False+False or True+True).

\subsection*{Reporting Methodology}

For higher confidence in the sampling procedure reported metrics, they are averages across $k$ runs on the entire test set, where $k=5$ for NCP and $k=10$ for Flawed Fictions (as the relatively smaller context length and inexpensive evaluation meant inference on Flawed Fictions was relatively inexpensive). $\pm \text{SEM}$ scores are run-level estimates of the uncertainty in the test-set mean metric, computed as the standard deviation of the $k$ run-level means divided by $\sqrt{k}$. This represents the run-level variability from LLM sampling.

Reported token counts include the generated latent tokens with the exception of during-training token counts in \autoref{tab:rl_gen_tokens}, as we did not log every reasoning trace produced during training. Note that COCONUT latent tokens are slightly computationally cheaper than those from CoLaR and LiteReason due to the lack of a latent head or Reasoning Projector. A potential alternative efficiency measurement could be to measure total FLOPs or another memory-utilization metric. We chose average tokens due to significant implementation differences that misrepresent true computational cost. For example, our method has been combined with VLLM which makes generation significantly more efficient than the other trained latent methods. 

Our best attempt at calculating the practical efficiency gains of LiteReason at test-time is presented in \autoref{tab:qwen25_wallclock}, where we compare the time-per-example between base, RL-Trained, and LiteReason models with and without latent reasoning. We find significant speedups with the LiteReason models, indicating that they are indeed more efficient in practice.

\begin{table*}[h]
{\small
\begin{tabular}{p{0.47\textwidth} p{0.47\textwidth}}
\begin{tcolorbox}[
colback=white,
  colframe=deeppurple,
  colbacktitle=deeppurple,
  coltitle=white,
  title = Reasoning \& Plan w/o Contrastive Reward,
  halign title=flush center,
]
 **Character Developments:**
 
   <citation>Source A says X</citation>, therefore <reasoning>Louise's motivations and emotions become more complex as the story progresses.</reasoning>
   
   <citation>Source A says X</citation>, therefore <reasoning>Louise begins to question her relationship with John Paul and her feelings for Lee Towson.</reasoning>
   \\ \noindent\rule{\textwidth}{0.4pt}
   In the next chapter, Louise's situation is revealed as she remains in a holding cell and is questioned by an officer. The chapter ends with the officer telling Louise of John Paul's assertion that she is involved in Barbara's disappearance. The chapter ends with Louise's thoughts and feelings about her relationship with John Paul and her own secrets. The chapter also concludes with the investigation into Barbara's disappearance and the revelation of John Paul's involvement. The chapter ends with Judy's decision to go home, and the chapter ends with the story near its conclusion. The chapter is marked by themes of family trauma, mental health, and survival, and the chapter ends with the discovery of Barbara's whereabouts. The chapter is an important chapter in the story, as it reveals the truth about Barbara's disappearance and the camp's dark past. The chapter ends with the story near its conclusion, and the chapter ends with the reader eager to know what will happen next. The chapter is a important chapter in the story, as it reveals the truth about Barbara's disappearance and the camp's dark past. The chapter ends with the reader eager to know what will happen next.  
\end{tcolorbox} & 
\begin{tcolorbox}[
colback=white,
  colframe=deeppurple,
  colbacktitle=deeppurple,
  coltitle=white,
  title = Reasoning \& Plan w/ Contrastive Reward,
  halign title=flush center, 
]
 **Revelation of John Paul's Claim:**
 
   <citation>Next chapter synopsis says the officer tells Louise of John Paul’s assertion that Louise is involved in Barbara’s disappearance.</citation> Therefore, the officer will reveal that John Paul has accused Louise of being involved in Barbara's disappearance, which will be a pivotal moment in the chapter. \\
   \noindent\rule{\textwidth}{0.4pt}
   Chapter 44 will begin with Louise being taken to a holding cell, where she will be interrogated by an officer who looks down on her. The officer will reveal that John Paul has accused Louise of being involved in Barbara's disappearance, which will be a significant moment of tension and conflict. Louise will respond defensively and sarcastically, using her wit and intelligence to try to deflect suspicion and protect herself. Throughout the scene, Louise will express her concern for her family, particularly her brother, Jesse, and her determination to protect him from their mother's neglect. The chapter will build on the complex relationships and motivations of the characters, particularly Louise, and set the stage for further revelations and developments in the story. 
\end{tcolorbox}
\end{tabular}
}
\caption{Example reasoning (snippet) and plans taken from models trained with and without the Contrastive Reward on the NCP task. We find that optimizing our contrastive reward produce are more specific to the given Story Information, and include less generic writing advice. The reasoning is also less formulaic, and obeys the prompt's format (with citation and reasoning tags) more faithfully.}
\label{tab:compare_plans_contrastive}
\end{table*}

\begin{table*}[t]
{\small 
\begin{tabular}{p{0.47\textwidth} p{0.47\textwidth}}

\begin{tcolorbox}[
  colback=white,
  colframe=deeppurple,
  colbacktitle=deeppurple,
  coltitle=white,
  halign title=flush center, 
  title=\textbf{Normal Prompt}
]

You are tasked with detecting the presence of continuity errors in a short story. A continuity error occurs when an event or detail in the story contradicts or is incompatible with previously established information about the story's world or characters. \\

Is there a continuity error in the provided story? Think step by step to answer this question. End your response with \textbackslash boxed\{Yes\} if you find a continuity error, and \textbackslash boxed\{No\} if you do not find a continuity error. If you respond \textbackslash boxed\{Yes\}, you should also provide the lines that introduce the continuity error and the lines from earlier in the story that are contradicted by the error. \\

Format these lines as follows:\par
<contradicted\_lines>\par
[If applicable, quote the lines from earlier in the story that are contradicted by the error]\par
</contradicted\_lines> \\

\textbackslash boxed\{answer\} \\

Here is the story to analyze: \\

<story> \\
\{story\} \\
</story> \\

Think carefully and check your work.

\end{tcolorbox}
& 

\begin{tcolorbox}[
  colback=white,
  colframe=deeppurple,
  colbacktitle=deeppurple,
  coltitle=white,
  halign title=flush center, 
  title=\textbf{Implicit Thought Prompt}
]
You are tasked with detecting the presence of continuity errors in a short story. A continuity error occurs when an event or detail in the story contradicts or is incompatible with previously established information about the story's world or characters. \\

Is there a continuity error in the provided story? Think step by step to answer this question. End your response with \textbackslash boxed\{Yes\} if you find a continuity error, and \textbackslash boxed\{No\} if you do not find a continuity error. If you respond \textbackslash boxed\{Yes\}, you should also provide the lines that introduce the continuity error and the lines from earlier in the story that are contradicted by the error. \\

Format these lines as follows:\par
<contradicted\_lines>\par
[If applicable, quote the lines from earlier in the story that are contradicted by the error]\par
</contradicted\_lines> \\

\textbackslash boxed\{answer\} \\

In addition to your normal reasoning, you may format your reasoning with implicit thought tags of the following format: \\

 <implicit\_thought>number</implicit\_thought> 
 
Where number is a number between 1 and 5, and corresponds to the complexity of the thought. When you use this tool, you can skip the sentence that you would have said and move on with your reasoning. This will allow you to think a lot more about the story, while skipping over obvious conclusions/reasoning steps. We recommend doing this in the center of your reasoning, and not at the beginning or end. \\

Here is the story to analyze: \\

<story> \\
\{story\} \\
</story> \\

Think carefully and check your work.

\end{tcolorbox}
\end{tabular}
\caption{Flawed Fictions normal and implicit-thought prompts adapted from the original provided by \citet{ahuja_finding_2025} to improve performance during RL.}
\label{tab:flawed_fiction_prompts}
}
\end{table*}

\begin{table*}[t]
{\small
\begin{tabular}{p{0.47\textwidth} p{0.47\textwidth}}
\begin{tcolorbox}[
colback=white,
  colframe=deeppurple,
  colbacktitle=deeppurple,
  coltitle=white,
  halign title=flush center, 
  title=\textbf{Thought Prompt}
]
 Instructions: You will be given the most recent chapter of the story, a high-level plan of the entire story, a summary of the previously written chapters, character sheets for the three main characters and a brief synopsis of what should happen in the next chapter. You will also be given a chapter header for the next chapter, containing the chapter's title and any other epigraph-type text. You will first reason about the given story and about what should come next. Next, you will write the next chapter of the story.

\#\#\# High-Level Story Summary/Plan: \#\#\#
\{global\_story\_sketch\}

\#\#\# Summary of Already Written Chapters: \#\#\#
\{previous\_chapter\_summaries\}

\#\#\# Character Sheets: \#\#\#
\{character\_sheets\}

\#\#\# Previous Chapter: \#\#\#
\{previous\_chapter\}

\#\#\# Next Chapter Synopsis: \#\#\#
\{next\_chapter\_synopsis\}

\#\#\# Next Chapter Header: \#\#\#
\{chapter\_header\}

\#\#\# Instructions: \#\#\#
Instructions: Based on the next chapter's synopsis and header, please reason step by step to come up with a more detailed plan for the next chapter. Format your reasoning with "<citation>source A says X</citation>, therefore <reasoning>reasoning</reasoning>" pairs, where the sources can be the character sheets, the high-level story plan, the previous-chapters summary, the next chapter synopsis, and the previous few chapters. Add and modify your conclusions as you remember more information. End your response with a detailed paragraph explaining your reasoning as to how the next chapter will unfold (including plot and character points), beginning this paragraph with "In summary: ".
\end{tcolorbox}
 &  
 \begin{tcolorbox}[
 colback=white,
  colframe=deeppurple,
  colbacktitle=deeppurple,
  coltitle=white,
  halign title=flush center, 
  title=\textbf{Implicit Thought Prompt}
]
 Instructions: You will be given the most recent chapter of the story, a high-level plan of the entire story, a summary of the previously written chapters, character sheets for the three main characters and a brief synopsis of what should happen in the next chapter. You will also be given a chapter header for the next chapter, containing the chapter's title and any other epigraph-type text. You will first reason about the given story and about what should come next. Next, you will write the next chapter of the story.

\#\#\# High-Level Story Summary/Plan: \#\#\#
\{global\_story\_sketch\}

\#\#\# Summary of Already Written Chapters: \#\#\#
\{previous\_chapter\_summaries\}

\#\#\# Character Sheets: \#\#\#
\{character\_sheets\}

\#\#\# Previous Chapter: \#\#\#
\{previous\_chapter\}

\#\#\# Next Chapter Synopsis: \#\#\#
\{next\_chapter\_synopsis\}

\#\#\# Next Chapter Header: \#\#\#
\{chapter\_header\}

\#\#\# Instructions: \#\#\#
Instructions: Based on the next chapter's synopsis and header, please reason step by step to come up with a more detailed plan for the next chapter. In addition to your normal reasoning, format your reasoning with two types of tags:
1) <citation>A says X</citation>, therefore <reasoning>B</reasoning> pairs, where the sources A can be the character sheets, the high-level story plan, the previous-chapters summary, the next chapter synopsis, and the previous few chapters. X is the relevant information within A, and B is your reasoning based on A and X.
2) <implicit\_thought>number</implicit\_thought> tags, where number is a number between 1 and 5, and corresponds to the complexity of the thought. When you use this tool, you can skip the sentence that you would have said and move on with your reasoning. This will allow you to think a lot more about the story, while skipping over obvious conclusions/reasoning steps. We recommend doing this in the center of your reasoning, and not at the beginning or end.
Add and modify your conclusions as you remember more information. End your response with a detailed paragraph explaining your reasoning as to how the next chapter will unfold (including plot and character points), beginning this paragraph with "In summary: ". 
\end{tcolorbox}
\end{tabular}
\caption{Next Chapter Prediction normal prompt and implicit-thought prompt. The prompts are almost exactly the same, but with an additional explanation for how to use the `implicit thought tags.'}
\label{tab:ncp_prompts}
}
\end{table*}
\section{Next Chapter Prediction: Human Evaluation}
\label{sec:ncp_human_evals}

We evaluate the chapters on the following dimensions:
(1)~\textbf{Plot:}  Does the chapter exhibit events and turns that move the plot forward logically?
(2)~\textbf{Creativity:}  Does it have engaging characters, themes, and imagery, and avoid overly cliched characters and storylines? (3)~\textbf{Development:} \  Does it introduce characters and settings with appropriate levels of detail and complexity? (4)~\textbf{Language Use:} \  Is the language varied and rich, exhibiting rhetorical, linguistic, and literary devices? (5)~\textbf{Characters:} Does it feature believable and conceptually consistent characters, including reasonable character arcs and development? (6)~\textbf{Overall Preference:}   Which of the two continuations did you prefer?

We generate chapters with Qwen2.5 7B-Instruct-1M. Relative strengths are shown in \autoref{fig:bradleyterryscores}. Across the majority of categories we find our LiteReason method to produce preferred chapters over the other baselines, best matching non-latent RL performance.

\subsection{Human Annotation Details}
\label{sec:annotation_details}

Annotators were recruited via Prolific, restricted to native English speakers employed in creative writing. 
In addition to the two test chapters, we show annotators the same Story Information as the models see, i.e.,   a global story sketch, previous chapter summaries, character sheets, the previous chapter, and a next chapter synopsis. 
Instructions were adapted from \citet{gurung_learning_2025}.  We paid annotators £14 per three datapoints or an estimated £9.33 per hour.

\end{document}